\documentclass[review]{elsarticle}
\usepackage[english]{babel}
\pdfoutput=1
\usepackage{amsmath}
\usepackage{enumitem}
\usepackage{subcaption}
\usepackage{placeins}
\graphicspath{ {./figures/} }
\usepackage{graphicx}
\usepackage[most]{tcolorbox}
\usepackage{caption}
\DeclareCaptionFormat{figrule}{#1#2#3\par\vspace{2pt}\noindent\rule{\linewidth}{0.4pt}}
\usepackage{multirow}
\usepackage{pdflscape}
\usepackage[colorlinks=true, allcolors=blue]{hyperref}
\usepackage{algorithm}
\usepackage{algorithmic}
\usepackage{threeparttable}
\usepackage{float}
\usepackage{verbatim} 
\usepackage{tabularx}
\usepackage{booktabs}
\usepackage{apalike}
\usepackage{longtable}
\usepackage{array}
\floatstyle{plaintop} 
\restylefloat{table}

\journal{Expert Systems with Applications}

\biboptions{authoryear}
\begin{document}
\begin{frontmatter}










\title{Ranking-Preserving Coreset Selection for Efficient Evolutionary Hyperparameter Optimization}

\author[nuces]{Mamoona Tasadduq\corref{cor1}}
\ead{l217809@lhr.nu.edu.pk}

\author[nuces]{Maryam Bashir}
\ead{maryam.bashir@nu.edu.pk}

\cortext[cor1]{Corresponding author}

\address[nuces]{National University of Computer and Emerging Sciences (FAST-NUCES), Lahore, Pakistan}


\begin{abstract}
Hyperparameter optimization (HPO) is essential for achieving high predictive performance, but its computational cost becomes prohibitive when each candidate configuration requires repeated model training on large datasets. This paper proposes a ranking-preserving coreset (RPC) selection framework for accelerating evolutionary HPO. Instead of selecting instances only to preserve predictive
accuracy, the proposed method constructs a compact coreset that preserves the relative ranking of hyperparameter configurations observed on the full training data. Coreset selection is formulated as an evolutionary optimization problem that maximizes Spearman rank correlation between full-data and subset-based configuration rankings while penalizing subset size. The selected coreset is subsequently used for hyperparameter optimization with three evolutionary algorithms: Genetic Algorithm (GA), Cross-generational elitist Selection, Heterogeneous Recombination and Cataclysmic Mutation (CHC), and Particle Swarm Optimization (PSO). Extensive experiments on seven tabular, two text, and two image benchmark datasets show that the proposed framework substantially reduces HPO time while maintaining test accuracy comparable to full-dataset HPO across Decision Tree, text classification, and image classification tasks. Across datasets and optimizers, the proposed method achieves HPO stage speedups ranging approximately from $1.14\times$ to $8.80\times$, while the one-time coreset construction cost is accounted for separately through end-to-end runtime and break-even analysis. On 20 held-out configurations not used during coreset construction, RPC achieved higher ranking correlation than same-size random sampling on 8 of 11 datasets (dataset-level Wilcoxon signed-rank test, \(p=0.014\)). This demonstrates that ranking preservation extends beyond the configurations used during selection. Coreset construction incurs a one-time cost, which can be amortized when the coreset is reused across HPO runs.

\end{abstract}


\end{frontmatter}

\section{Introduction}
\label{sec:intro}
Machine learning has revolutionized our ability to extract meaningful insights from vast and diverse datasets across a wide range of application domains. Among its various paradigms, deep learning has gained significant attention for its ability to automatically learn hierarchical representations from data. Machine learning and deep learning models have demonstrated remarkable success in areas such as classification, prediction, computer vision, and natural language processing. However, training and optimizing these models often require substantial computational resources, particularly when dealing with large, high-dimensional datasets. The selection of hyperparameters often influences model performance and accuracy~\citep{30}. Depending on the learning algorithm, these may include parameters such as learning rate, batch size, optimizer settings, tree depth, network architecture, splitting criteria, and regularization parameters. Identifying suitable hyperparameter configurations is essential for achieving optimal predictive performance. Nevertheless, balancing computational efficiency and model performance remains a significant challenge, especially when the size, complexity, and dimensionality of modern datasets continue to grow.

The computational burden of training and optimizing machine learning and deep learning models can be reduced through techniques such as instance selection, feature selection, and efficient hyperparameter optimization~\citep {31,32}. Among these, instance selection aims to identify a representative subset of instances from the original dataset while preserving its essential characteristics. Instance selection has proven effective at reducing computational costs by decreasing the amount of data processed during model training and evaluation \citep{20}. In addition, hyperparameter optimization seeks to identify the combination of hyperparameter values within a predefined search space that yields the best predictive performance \citep{27}. However, repeatedly evaluating numerous hyperparameter configurations remains computationally expensive, particularly for large-scale and high-dimensional datasets.

Hyperparameter optimization has become a fundamental component of modern machine learning pipelines, as model performance is often highly sensitive to hyperparameter choices \citep{30}. While traditional hyperparameter optimization techniques such as Grid Search and Random Search are valued for their straightforward implementation, they scale poorly and suffer from prohibitive computational overhead as the dimensionality of the search space increases \citep{1, 2, 27}. 
To mitigate these computational bottlenecks, more sophisticated sequential methodologies, such as Bayesian Optimization, Hyperband, and metaheuristic evolutionary algorithms, have been proposed to improve search efficiency while reducing the number of costly model evaluations \citep{3, 4, 6}. Consequently, modern frameworks like Optuna have been deployed to optimize complex architectures, such as tuning an XGBoost classifier for cardiovascular disease prediction, without the computational overhead of traditional search methods \citep{6}. Despite these advances, the repeated training and evaluation of models across numerous candidate configurations remain a major computational bottleneck, particularly for large-scale datasets and deep learning applications. For instance, training 10 candidates over 10 generations entails approximately 100 model training runs per HPO run. Repeated search across multiple optimizers and seeds can multiply this cost, particularly when using the full training dataset. Coreset construction incurs an upfront overhead; therefore, its overall runtime benefit only comes when the time saved across HPO runs exceeds the time spent constructing the coreset.
 
Recent advances in dataset reduction and coreset selection have demonstrated that carefully selected subsets can substantially reduce the computational cost of model training while preserving predictive performance \citep{20}. Gradient-based approaches such as GLISTER \citep{22} and GRAD-MATCH \citep{21}  select training subsets by optimizing validation performance and matching full-data gradients, respectively, whereas more recent methods have explored clustering-based \citep{23}, bi-level optimization \citep{25}, and submodular \citep{24} strategies for efficient data selection in both conventional machine learning and large-scale language models. Despite their effectiveness in reducing training costs, these approaches primarily focus on preserving model performance and training dynamics rather than preserving the relative ranking of hyperparameter configurations. Consequently, a subset that performs well during model training may not preserve the relative ranking of hyperparameter configurations needed for reliable hyperparameter optimization. This limitation motivates the development of ranking-aware data-reduction strategies specifically designed to accelerate hyperparameter optimization. Ranking-aware surrogate methods have investigated relative
configuration ordering as an alternative to predicting exact performance values. 
For instance, \cite{26} proposed Deep Ranking Ensembles, which use ranking-based surrogate models to guide HPO by prioritizing the relative ordering of candidate configurations
rather than precise performance predictions. This motivates
investigating ranking agreement as a criterion for training-subset
selection. Whereas Deep Ranking Ensembles learn a surrogate to
guide configuration selection, RPC selects the training instances
used to evaluate configurations.

Ranking preservation is relevant to evolutionary HPO because
selection decisions often depend on the relative performance of
candidate configurations. If a subset preserves their ordering,
rank-based decisions, such as tournament selection and elite
retention, can remain unchanged despite differences in absolute
validation scores. However, ranking agreement on reference
configurations does not guarantee an identical search trajectory
or final predictive performance: unseen configurations, ties,
and decision rules involving score magnitudes can still affect
the search.

Motivated by these considerations, we propose a
ranking-preserving coreset framework that optimizes agreement
between full-data and subset-induced hyperparameter rankings
over reference configurations while penalizing subset size. Unlike conventional instance selection and coreset methods that focus primarily on preserving predictive accuracy or training dynamics, the proposed method preserves the configuration ranking structure that guides evolutionary HPO algorithms, enabling them to identify promising regions of the search space without requiring exact replication of full-dataset validation accuracies. This makes the proposed framework suitable for repeated model-selection workflows, where high-quality configurations must be identified with fewer evaluations on the full training dataset.

Based on the identified research gap, this study addresses the following research questions:

\begin{itemize}
    \item \textbf{RQ1:} Can a compact ranking-preserving coreset maintain predictive performance close to full-dataset hyperparameter optimization while using substantially fewer training instances?

    \item \textbf{RQ2:} How effectively does the proposed coreset preserve the relative ranking of hyperparameter configurations compared with same-size random and gradient-based subsets?

    \item \textbf{RQ3:} Does the ranking preservation achieved during coreset construction generalize to unseen held-out hyperparameter configurations?

    \item \textbf{RQ4:} Under what conditions does the proposed framework provide computational savings after accounting for the one-time coreset construction cost?
\end{itemize}

\section*{Main Contributions}
The primary contributions of this work are summarized as follows:
\begin{enumerate}

    \item \textbf{Ranking-Preserving Coreset Selection Framework:} 
    We propose a ranking-preserving coreset selection framework that constructs compact training subsets specifically designed to preserve the relative performance ordering of hyperparameter configurations observed on the full training dataset.
    
    \item \textbf{Evolutionary Ranking-Aware Optimization Objective:} 
    We formulate coreset selection as an evolutionary optimization problem that jointly maximizes ranking fidelity, measured using Spearman rank correlation, and penalizes subset size. This enables the construction of compact coresets that preserve the relative ranking of hyperparameter configurations.
    
    \item \textbf{Efficient Evolutionary Hyperparameter Optimization:} 
    We integrate the selected coreset into evolutionary HPO, allowing GA, CHC, and PSO to search for high-quality hyperparameter configurations using substantially fewer training instances while maintaining performance close to that of full-dataset HPO.
    
    \item \textbf{Comprehensive Multi-Modal Evaluation:} 
    We evaluate the proposed framework across 11 benchmark datasets, including seven UCI tabular datasets, two text datasets (IMDb and TweetEval Sentiment), and two image datasets (CIFAR-10 and DermaMNIST). The evaluation covers Decision Trees for tabular data, TF-IDF with a PyTorch MLP for text classification, and ResNet18 for image classification.

\end{enumerate}

The rest of this paper is organized as follows. Section~\ref{sec:related_work} reviews the literature on instance selection, coreset construction, and hyperparameter optimization. Section~\ref{sec:methodology} presents the proposed ranking-preserving coreset selection framework and the evolutionary optimization strategy. Section~\ref{sec:experiments} describes the experimental setup. Section~\ref{sec:results_discussion} reports and discusses the results obtained on UCI tabular datasets, text classification datasets, and image classification datasets. Finally, Section~\ref{sec:conclusion} concludes the paper and outlines directions for future research.

\section{Related Work}
\label{sec:related_work}

In this section, we present a comprehensive review of the latest studies related to instance selection, coreset construction, hyperparameter optimization, and evolutionary search. The aim is to explore the methods that reduce computational cost through representative data selection and efficient hyperparameter exploration, as these areas are most closely related to the proposed ranking-preserving coreset selection framework. Instance selection aims to identify a representative subset of data instances that preserves the essential characteristics of the original dataset while reducing computational cost. Because subset selection can be formulated as a search problem, evolutionary and metaheuristic approaches have been widely investigated for this purpose.

Various studies have explored the potential of using evolutionary algorithms for instance selection in different domains. \cite{7} investigated the potential of genetic algorithms for instance and feature selection in knowledge-based systems to enhance computational efficiency. Another paper authored by \cite{8} proposes a framework that uses fuzzy clustering decomposition in combination with a genetic algorithm. The framework uses fuzzy clustering to decompose the dataset into coherent clusters and selects the representative instances from these clusters to perform regression tasks. The authors underscore the effectiveness of this hybrid model for high-dimensional data in improving model performance and generalization for regression problems. The study conducted by \cite{9} evaluates the effectiveness of fifteen instance selection methods within the domain of text sentiment classification. The authors highlight the performance of different instance selection methods for data reduction, providing insights into the appropriate selection of an instance selection technique for text sentiment analysis. \cite{10} proposed a Multi-Objective CHC evolutionary algorithm for instance selection that integrates the non-dominated sorting with CHC. The proposed algorithm demonstrated 50\% reduction in the dataset while not compromising data accuracy for most of the dataset when compared with KNN.  

Recent advances in coreset construction have further demonstrated the effectiveness of selecting representative subsets in reducing computational cost while preserving model performance. Gradient-based approaches such as GRAD-MATCH \citep{21} and GLISTER \citep{22} construct representative subsets by matching gradient information or optimizing validation performance through bilevel optimization. Other studies have explored clustering-based \citep{23}, submodular \citep{24}, and bilevel optimization strategies \citep{25} for efficient data reduction in both conventional machine learning and large-scale language models. In parallel, recent work by \cite{26} highlighted the importance of preserving the relative ranking of hyperparameter configurations during optimization, demonstrating that ranking-aware surrogate models can effectively guide hyperparameter search without requiring highly accurate performance prediction. While these studies illustrate the importance of both representative data selection and ranking preservation, they have largely been investigated independently. 
The methods compared in  Table~\ref{tab:coreset_comparison} differ in their optimization targets: existing subset-selection methods emphasize gradients, validation performance, or data
representativeness, whereas ranking-aware surrogates guide configuration selection. RPC instead directly optimizes
configuration-ranking agreement during training-subset selection.

In the literature, many studies have explored the importance and techniques of hyperparameter tuning for deep learning tasks.  The use of genetic algorithms for HPO has gained significant attention in the domain of machine learning and optimization. Researchers have explored the potential of GAs to enhance the efficiency and effectiveness of hyperparameter optimization. However, a comparative study conducted by \cite{11} highlights the strengths and limitations of three hyperparameter techniques, i.e., Bayesian, grid, and GA. The authors conducted empirical evaluations and provided insights into the trade-offs and capabilities of each optimization method for a particular scenario.   

Evolutionary HPO studies provide methodological context for the GA and PSO optimizers evaluated in this work. They investigate how evolutionary search identifies promising hyperparameter configurations, while RPC addresses the training-data cost of evaluating those configurations. \cite{12} demonstrated the applicability of GAs in HPO for deep learning models in the context of PM2.5 time-series prediction, emphasizing their potential in improving predictive accuracy. It showed promising results when compared with default hyperparameters and random search algorithms for air quality forecasting. The proposed HPO outperformed other methods by significantly reducing the MSE. 
Additionally, \cite{13} demonstrated the use of PSO and GA to select an optimal set of hyperparameters for machine learning models used in high-energy physics data analysis, achieving significantly improved sensitivity (by 12-13\%) compared with the default hyperparameter settings. The results of another study showed the effectiveness of GA in optimizing hyperparameters for predictive business process monitoring \citep{14}. \cite{15} proposed a qualitative approximation approach for feature selection using a genetic algorithm. This study aims to identify the best feature subset from the complex dataset more efficiently, thereby offering a lightweight model to enhance performance. \cite{16} leverage GA to jointly optimize a neural network’s architecture, such as the number of neurons and layers, and hyperparameters, such as the learning rate and batch size. The empirical results of the study demonstrate effectiveness in performance, thereby providing insights for enhancing the capabilities of deep learning models. 

Hyperparameter optimization is equally important in deep CNNs, where complex, high-dimensional data, particularly images, are used for analysis and insight generation. \cite{17} addresses the challenges of hyperparameter tuning to enhance the performance of image-related tasks. The authors explored a genetic algorithm-based approach to enhance the performance of convolutional neural networks.  In another study, \cite{18} uses a genetic algorithm to select the hyperparameters for a CNN model, used for bone age assessment and classification. The study demonstrates the importance of integrating hyperparameter optimization into advanced machine learning techniques to improve the efficiency of bone age assessment in medical imaging. 

Recent studies also investigate the efficiency of configuration
search. \citep{35} compared Grid, Random, and
Bayesian search for heart-failure prediction, reporting lower processing time for Bayesian Search in their evaluated setting. \citep{34} introduced a multi-strategy optimizer that modifies initialization and search updates to improve convergence and applied it to ResNet18 hyperparameter tuning.
These studies address the configuration-search component, whereas RPC reduces the training data used for configuration evaluation through a ranking-preserving subset-selection objective. 

While these approaches improve search efficiency, their computational cost remains largely dependent on the repeated training and evaluation of models across candidate configurations. This challenge becomes particularly pronounced for large-scale datasets and deep learning applications, motivating the use of representative data subsets to accelerate optimization.

The reviewed literature demonstrates that evolutionary algorithms are effective for both instance selection and HPO, while recent coreset construction techniques offer promising mechanisms to reduce training costs through representative data selection. However, existing studies primarily focus on preserving predictive performance, gradient information, or training dynamics. Similarly, ranking-based hyperparameter optimization methods focus on learning ranking-aware surrogate models rather than constructing representative data subsets. The reviewed subset selection methods mainly focus on predictive performance, gradients, or data representativeness, leaving configuration-ranking preservation as a direction to explore.
This observation motivates the present work, which investigates whether ranking-preserving selection can be used to construct compact coresets that retain the hyperparameter landscape of the original dataset while substantially reducing the computational cost of HPO using evolutionary algorithms.

\subsection{Comparison with Existing Coreset Selection Methods}

Existing methods on coreset or instance selection are generally designed to reduce computational cost while preserving predictive performance, data distribution, or gradient information. Gradient-matching methods such as GRAD-MATCH \citep{21} aim to select subsets whose gradients approximate those of the full training dataset. GLISTER \citep{22} formulates subset selection as a bi-level optimization problem that selects training examples expected to improve validation performance. AutoMata \citep{5} extends gradient-based subset
selection to the setting of compute-efficient hyperparameter tuning. Other approaches are based on clustering, submodular maximization, or diversity-based selection to select representative samples from the original dataset~\citep{23,33,24}.

Although these methods are effective at selecting a subset to reduce model training costs, they do not directly optimize the agreement between full-data and subset-induced configuration rankings. In HPO, the central requirement is not to reproduce the exact validation accuracy of every candidate configuration but to preserve the hyperparameter landscape. An HPO algorithm primarily needs to identify which configurations are better than the others so that the search process can move toward promising regions of the hyperparameter space. Therefore, a subset that preserves predictive accuracy or gradient similarity may still distort the ranking of candidate hyperparameter configurations, thereby misleading the optimizer.

The proposed method differs from existing coreset selection strategies by directly optimizing the ranking fidelity of the hyperparameter space. Instead of selecting instances to approximate the gradients, the full-data loss, or the class distribution, the proposed framework uses an evolutionary process to select a subset that maximizes the Spearman rank correlation between the full-data and subset-based hyperparameter rankings while penalizing larger subsets. This objective is specifically designed for HPO because it preserves the relative ordering of candidate configurations, which evolutionary optimizers use during selection, recombination, and update operations.

The theoretical advantage of the proposed objective is its alignment with the order-based nature of hyperparameter optimization. Suppose $R_D$ denotes the ranking of candidate configurations obtained from the full training dataset, and let $R_S$ denote the ranking from a selected subset. If $R_S$ closely matches $R_D$, then the subset preserves the preference structure among configurations. Consequently, configurations that are promising on the full dataset are also likely to remain promising on the subset. Unlike accuracy-preserving objectives, rank-based objectives are invariant to strictly increasing transformations of validation scores and do not require the subset to reproduce exact performance values. This motivates ranking agreement as a criterion for selecting training subsets for HPO.  Table~\ref{tab:coreset_comparison} summarizes the main differences between existing coreset methods and the proposed ranking-preserving coreset selection framework.

\begin{landscape}
\begin{table}[h]
\footnotesize
\caption{Comparison of existing coreset selection methods with the proposed 
ranking-preserving coreset framework.}
\label{tab:coreset_comparison}
\begin{tabular}{p{4.5cm}p{4.0cm}p{4.0cm}p{4.5cm}}
\hline
\textbf{Method Category} & \textbf{Main Optimization Objective} & \textbf{Primary Use Case} & \textbf{Relevance to HPO} \\
\hline
Gradient-matching methods, e.g., GRAD-MATCH  \citep{21}
& Match subset gradients to full-data gradients
& Efficient model training
& Preserving gradients does not necessarily preserve the ranking of hyperparameter configurations. \\
Validation-aware methods, e.g., GLISTER \citep{22}
& Select examples that improve validation performance through bilevel optimization
& Generalization-oriented subset selection
& Optimizes predictive performance rather than the hyperparameter ranking landscape. \\
HPO-oriented gradient subset methods, e.g., AutoMata \citep{5}
& Reduce compute cost for hyperparameter tuning using gradient-based subset selection
& Compute-efficient HPO
& Focuses on gradient/accuracy preservation and does not explicitly optimize rank correlation among configurations. \\
Clustering/diversity/ submodular methods \citep{23,33,24}
& Preserve data distribution, diversity, or representative samples
& Dataset summarization and training acceleration
& Representative data subsets may still distort the ordering of candidate hyperparameter configurations. \\
Ranking-aware surrogate methods, e.g., Deep Ranking Ensembles~\citep{26}
& Learn the relative ordering of configuration performance
using learning-to-rank objectives
& Surrogate-guided hyperparameter optimization
& Predicts configuration rankings to guide the search;
does not select a training subset to preserve full-data rankings. \\

\textbf{Proposed method}
& \textbf{Maximize Spearman rank correlation between full-data and subset-based configuration rankings while penalizing subset size}
& \textbf{Ranking-preserving coreset selection for evolutionary HPO}
& \textbf{Directly optimizes ranking agreement through
training-subset selection; generalization beyond construction
anchors is evaluated on held-out configurations.} \\

\hline
\end{tabular}
\end{table}
\end{landscape}

\section{Methodology}
\label{sec:methodology}

This section presents the proposed ranking-preserving coreset selection framework for efficient hyperparameter optimization. The framework consists of two stages: (i) an evolutionary coreset selection stage that constructs a compact representative coreset while preserving the hyperparameter ranking structure of the original dataset, and (ii) an evolutionary hyperparameter optimization stage performed on the selected coreset. The overall objective is to reduce computational cost while maintaining the ability to identify high-quality hyperparameter configurations.

\subsection{Overview of the Proposed Framework}

Let $D_{train}$ denote the full training dataset containing $N$ instances. The proposed framework first evaluates a set of candidate hyperparameter configurations on the full training dataset to obtain a reference ranking. An evolutionary coreset selection procedure is then employed to search for candidate subsets $S \subset D_{train}$ and identify a compact subset $S^* \subset D_{train}$ that preserves this ranking as closely as possible while minimizing its size. Finally, hyperparameter optimization is performed on the selected coreset $S^*$ using three different evolutionary algorithms. The same validation set is used to compute the reference ranking during coreset selection and to evaluate candidate configurations during HPO, while the test set is kept completely unseen and used only for final performance reporting.

\begin{figure}[H]
\includegraphics[scale=0.45]
{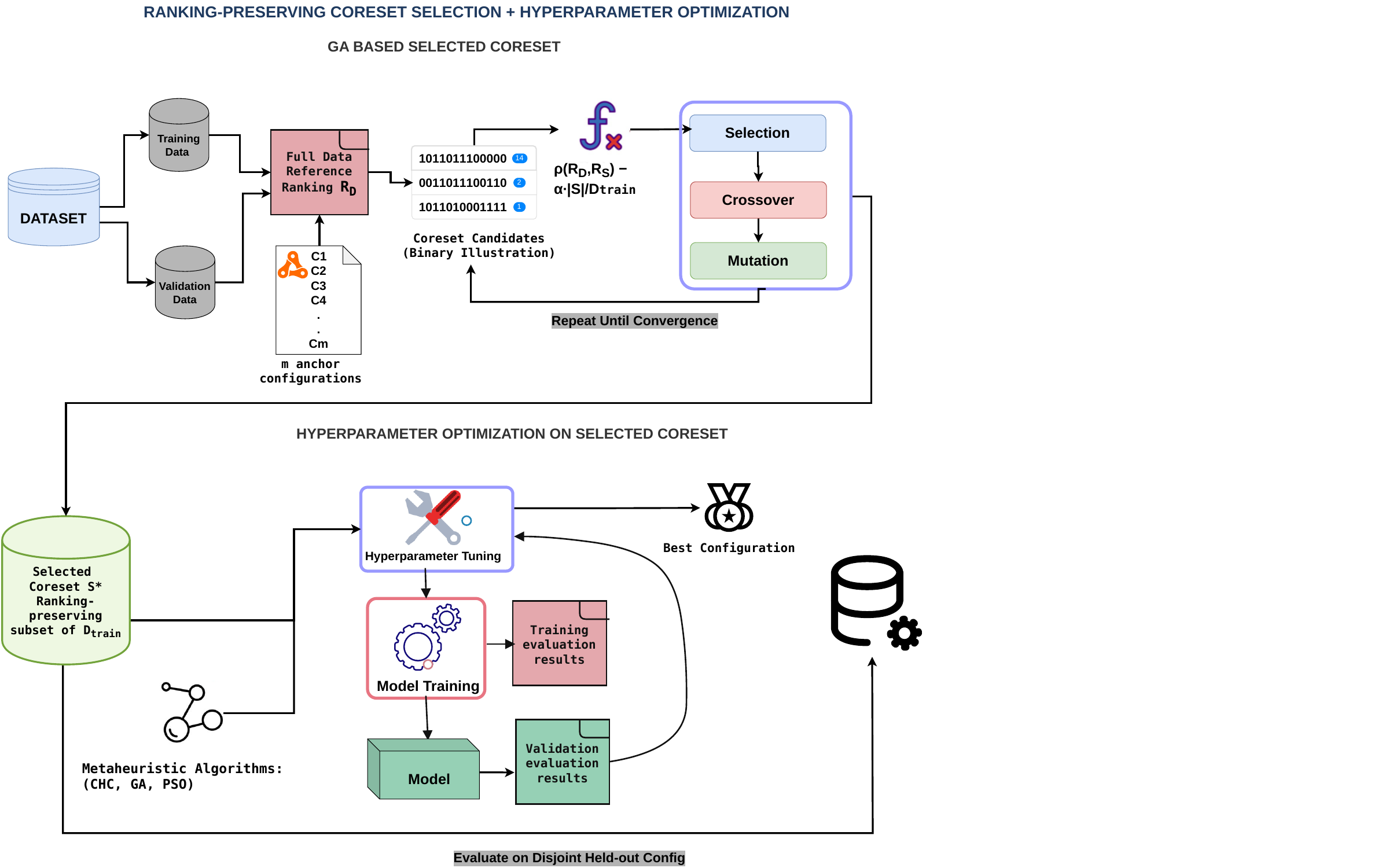}
\caption{
Overview of the proposed ranking-preserving coreset selection framework. A reference ranking of hyperparameter configurations is first obtained from the full training dataset. An evolutionary instance selection procedure then constructs a compact coreset by maximizing the Spearman rank correlation between the coreset and full-dataset rankings while minimizing subset size. Hyperparameter optimization is subsequently performed on the selected coreset using GA, CHC, or PSO, reducing computational cost using the selected subset for configuration evaluation.
}
\label{fig:archi}
\end{figure}

Figure~\ref{fig:archi} illustrates the overall workflow of the proposed framework. Since the coreset selection stage is executed only once for a given dataset, the resulting coreset can be reused across multiple hyperparameter optimization runs, thereby amortizing the sampling cost over time.

\subsection{Ranking-Preserving Coreset Selection}

The central hypothesis of this work is that hyperparameter optimization primarily depends on preserving the relative ordering of candidate configurations rather than accurately reproducing their exact performance values.

Let

\begin{equation}
C=\{c_1,c_2,\ldots,c_m\}
\end{equation}

denotes a set of sampled hyperparameter configurations.

Each configuration is evaluated on the full training dataset $D_{train}$ using a performance function $f(\cdot)$, producing

\begin{equation}
P_D=
\left[
f(c_1,D_{train},D_{val}),
f(c_2,D_{train},D_{val}),
\ldots,
f(c_m,D_{train},D_{val})
\right].
\end{equation}

The corresponding reference ranking is defined as

\begin{equation}
R_D=\text{rank}(P_D).
\end{equation}

The objective of coreset selection is to identify a subset

\begin{equation}
S \subset D_{train}
\end{equation}

such that the ranking induced by $S$ closely approximates the ranking obtained from the full training dataset.

For a candidate subset $S$, the same configurations are evaluated to obtain

\begin{equation}
P_S=
\left[
f(c_1,S,D_{val}),
f(c_2,S,D_{val}),
\ldots,
f(c_m,S,D_{val})
\right].
\end{equation}

and

\begin{equation}
R_S=\text{rank}(P_S).
\end{equation}

The similarity between $R_D$ and $R_S$ is measured using the Spearman rank correlation coefficient:

\begin{equation}
\rho(R_D,R_S)=\operatorname{Corr}(R_D,R_S),
\end{equation}
where $\operatorname{Corr}$ denotes the normalized covariance of the two rank vectors. Tied scores are assigned average ranks. A value of $\rho=1$ indicates perfect ranking preservation, whereas lower values indicate increasing deviation from the reference ranking.



\subsection{Ranking-Based Approximation}

The proposed framework seeks to preserve the relative ordering of hyperparameter configurations observed on the full training dataset. Rather than reproducing the exact validation performance values of the anchor configurations, the objective is to preserve their ranking structure. Accordingly, Spearman rank correlation is used to quantify the agreement between the rankings obtained from the full training data and from a candidate subset. A high Spearman correlation therefore indicates that the subset preserves the relative ordering of the evaluated hyperparameter configurations.

To simultaneously maximize ranking fidelity and minimize subset size, the fitness function is defined as

\begin{equation}
\mathrm{Fitness}(S)
=
\rho(R_D,R_S)
-
\alpha
{|S|} 
/ 
{|D_{train}|}
\label{eq:fitness}
\end{equation}

where $\alpha$ controls the trade-off between ranking preservation and subset compactness. We use $\alpha=0.20$ as a fixed heuristic setting rather than an experimentally established optimum.
For the permitted subset fractions of 0.10--0.30, this produces a size penalty of 0.02--0.06. Increasing $\alpha$ strengthens the preference for smaller subsets, whereas decreasing it gives greater relative weight to anchor ranking agreement.

The evolutionary search process iteratively improves candidate subsets using selection, crossover, and mutation operators. The subset achieving the highest fitness value is selected as the final coreset $S^*$.

Each chromosome represents a candidate coreset through its selected training-instance indices. This is equivalent to a binary membership representation, where 1 indicates a selected
instance and 0 an excluded instance. Stratified initialization is used to preserve class coverage in the initial population. Parent chromosomes are selected using tournament selection, offspring are generated using union-based crossover, and mutation is performed by randomly adding or removing instances. A repair step ensures that the selected subset remains within the predefined fraction range (0.1 to 0.3) and maintains class coverage. Elitism is applied by directly retaining the top-E chromosome(s) from the current population, while the rest of the population is generated through tournament selection, crossover, and mutation. The search terminates when the maximum number of generations is reached or when no improvement is observed for $\pi$ consecutive generations. The details are presented in Algorithm ~\ref{alg:active_sampling}.
\vspace{1em}

\begingroup
\setlength{\abovedisplayskip}{2pt}
\setlength{\belowdisplayskip}{2pt}
\setlength{\abovedisplayshortskip}{2pt}
\setlength{\belowdisplayshortskip}{2pt}

\noindent
\begin{tcolorbox}[
breakable,
enhanced,
colback=white,
colframe=black,
boxrule=0.4pt,
arc=0.5mm,
boxsep=0.5pt,
left=1mm,
right=1mm,
top=0mm,
bottom=0mm,
before skip=0pt,
after skip=0pt,
parbox=false
]

\captionof{algorithm}{Ranking-Preserving Coreset Selection}
\label{alg:active_sampling}

\scriptsize

\begin{algorithmic}[1]
\setlength{\itemsep}{0pt}
\setlength{\parskip}{0pt}
\setlength{\parsep}{0pt}

\STATE \textbf{Input:} Training set $D_{train}=\{(x_i,y_i)\}_{i=1}^{N}$, validation set $D_{val}$, hyperparameter search space $\mathcal{H}$, number of anchor configurations $m$, candidate pool size $M$, coreset population size $P_S$, maximum generations $G_S$, subset fraction bounds $[f_{\min},f_{\max}]$, mutation rate $p_m$, crossover probability $p_c$, tournament size $\tau$, size penalty $\alpha$, patience $\pi$

\STATE \textbf{Experimental setting:} In the main experiments, we use $m=20$ anchor configurations, elite count $E=1$, mutation rate $p_m=0.10$, crossover probability $p_c=1.0$, i.e., crossover is applied to every generated child, size penalty $\alpha=0.20$, and subset fraction bounds $[0.1,0.3]$.

\STATE \textbf{Output:} Ranking-preserving coreset $S^*$

\STATE \textbf{Encoding:} The coreset-search population is 
$\mathcal{P}_S=\{z_{1},z_{2},\ldots,z_{P_S}\}$, where each chromosome $z_{r}$ represents one candidate coreset and is encoded as a set of selected training indices. The subset represented by $z_{r}$ is 
$S(z_{r})=\{(x_i,y_i)\in D_{train}: i\in z_{r}\}$.

\STATE Sample a candidate configuration pool $\widetilde{C}=\{\tilde{c}_1,\tilde{c}_2,\ldots,\tilde{c}_M\}$ from $\mathcal{H}$.

\STATE Select $m$ diverse anchor configurations $C=\{c_1,c_2,\ldots,c_m\}$ from $\widetilde{C}$ using maximin farthest-point sampling, where each newly selected anchor maximizes its minimum distance to the already selected anchors.

\STATE Train and evaluate each anchor configuration $c_j \in C$ on the full training set $D_{train}$ using $D_{val}$.

\STATE Obtain the full-data validation performance vector:
\[
P_D=[f(c_1,D_{train},D_{val}),\ldots,f(c_m,D_{train},D_{val})].
\]

\STATE Compute the reference full-data ranking:
\[
R_D=\operatorname{rank}(P_D).
\]

\STATE Initialize coreset population $\mathcal{P}_S$ with $P_S$ chromosomes using stratified random subset sampling. Each chromosome is assigned a random subset fraction $f \in [f_{\min},f_{\max}]$.

\STATE Repair each chromosome to preserve class coverage and satisfy:
\[
f_{\min} \leq |S(z)|/N \leq f_{\max}.
\]

\STATE Set $S^*=\emptyset$, $best\_fitness=-\infty$, and $no\_improve=0$.

\FOR{$g=1$ to $G_S$}

    \STATE Initialize evaluated population $\mathcal{E}=\emptyset$.

    \FOR{each chromosome $z_i \in \mathcal{P}_S$}

        \STATE Evaluate all anchor configurations $c_j \in C$ on subset $S(z_i)$ using $D_{val}$.

        \STATE Obtain subset performance vector:
        \[
        P_{S_i}=[f(c_1,S(z_i),D_{val}),\ldots,f(c_m,S(z_i),D_{val})].
        \]

        \STATE Compute subset-induced ranking:
        \[
        R_{S_i}=\operatorname{rank}(P_{S_i}).
        \]

        \STATE Compute ranking-preservation fitness:
        \[
        F(z_i)=\rho(R_D,R_{S_i})-\alpha\frac{|S(z_i)|}{|D_{train}|}.
        \]

        \STATE Store $(z_i,F(z_i))$ in $\mathcal{E}$.

    \ENDFOR

    \STATE Sort $\mathcal{E}$ in descending order of fitness.

    \STATE Let $z_{\text{best}}$ be the best chromosome in the current generation.

    \IF{$F(z_{\text{best}})>best\_fitness$}
        \STATE Update $S^*=S(z_{\text{best}})$ and $best\_fitness=F(z_{\text{best}})$.
        \STATE Set $no\_improve=0$.
    \ELSE
        \STATE Set $no\_improve=no\_improve+1$.
    \ENDIF

    \IF{$\rho_{\text{best}} \geq 0.9999$ and $agreement_{\text{best}} \geq 0.9999$}
        \STATE Stop early because perfect ranking preservation is reached.
        \STATE \textbf{break}
    \ENDIF

    \IF{$no\_improve \geq \pi$}
        \STATE Stop early due to no fitness improvement.
        \STATE \textbf{break}
    \ENDIF

    \STATE Initialize next population $\mathcal{P}_{next}$ with the top-$E$ elite chromosomes from $\mathcal{E}$.

    \WHILE{$|\mathcal{P}_{next}| < P_S$}

        \STATE Select parent chromosomes $z_a$ and $z_b$ using tournament selection of size $\tau$.

        \STATE Randomly sample a target subset fraction $f \in [f_{\min},f_{\max}]$.

        \STATE Generate child chromosome $z_o$ using subset crossover. Specifically, form the union of selected indices from both parents:
        \[
        U=z_a \cup z_b.
        \]

        \STATE Sample candidate selected indices from $U$ to form $z_o$.

        \STATE Apply class-wise mutation to $z_o$ using mutation rate $p_m$. Within each class, remove a small proportion of selected instances and replace them with other instances from the same class.

        \STATE Repair $z_o$ to preserve class coverage and satisfy:
        \[
        f_{\min} \leq {|S(z_o)|}/{N} \leq f_{\max}.
        \]
        \STATE Add $z_o$ to $\mathcal{P}_{next}$.
    \ENDWHILE

    \STATE Replace the current coreset population:
    \[
    \mathcal{P}_S \leftarrow \mathcal{P}_{next}.
    \]

\ENDFOR

\STATE Return final ranking-preserving coreset $S^*$.

\end{algorithmic}
\end{tcolorbox}
\endgroup
\vspace{1em}

\subsection{Evolutionary Hyperparameter Optimization}

Once the coreset $S^*$ has been selected, hyperparameter optimization is performed using evolutionary search.

Let $\Theta$ denote the hyperparameter search space, and let $\theta \in \Theta$ denote a candidate hyperparameter configuration. Each individual represents a candidate hyperparameter configuration $\theta_i$. The optimization objective is

\begin{equation}
\theta^*
=
\arg\max_{\theta\in\Theta}
f(\theta,S^*,D_{\mathrm{val}}),
\label{eq:hpo_objective}
\end{equation}

where $f(\theta,S^*,D_{\mathrm{val}})$ denotes the validation
performance obtained by training configuration $\theta$ on the
selected coreset $S^*$.
The complete evolutionary HPO procedure is presented in Algorithm~\ref{alg:hpo}.

Three evolutionary optimization techniques are investigated:

\begin{enumerate}[noitemsep]
    \item Genetic Algorithm (GA)
    \item Cross-generational elitist selection, Heterogeneous recombination, and Cataclysmic mutation (CHC)
    \item Particle Swarm Optimization (PSO)
\end{enumerate}

These algorithms employ different exploration and exploitation strategies, enabling a comprehensive evaluation of the proposed framework.

\vspace{2em}
\begingroup
\setlength{\abovedisplayskip}{2pt}
\setlength{\belowdisplayskip}{2pt}
\setlength{\abovedisplayshortskip}{2pt}
\setlength{\belowdisplayshortskip}{2pt}

\noindent
\begin{tcolorbox}[
breakable,
enhanced,
colback=white,
colframe=black,
boxrule=0.4pt,
arc=0.5mm,
boxsep=0.5pt,
left=1mm,
right=1mm,
top=0mm,
bottom=0mm,
before skip=0pt,
after skip=0pt,
parbox=false
]

\captionof{algorithm}{Evolutionary Hyperparameter Optimization}
\label{alg:hpo}

\scriptsize

\begin{algorithmic}[1]
\setlength{\itemsep}{0pt}
\setlength{\parskip}{0pt}
\setlength{\parsep}{0pt}

\STATE \textbf{Input:}
Coreset $S^*$,
search space $\Theta$,
optimizer $\mathcal{O}$

\STATE Initialize population of candidate
hyperparameter configurations
\[
H=\{\theta_1,\theta_2,\ldots,\theta_n\}
\]
\FOR{each generation}
    \FOR{each configuration $\theta_i$}
        \STATE Train model using $S^*$
        \STATE Evaluate performance on $D_{val}$
        \[
        f(\theta_i)=Acc(\theta_i,S^*,D_{val})
        \]
    \ENDFOR
    \IF{$\mathcal{O}$ = GA}
        \STATE Apply selection, crossover,
        and mutation
    \ELSIF{$\mathcal{O}$ = CHC}
        \STATE Apply HUX recombination,
        elitist selection,
        and cataclysmic mutation
    \ELSIF{$\mathcal{O}$ = PSO}
        \STATE Update particle velocity
        and position
    \ENDIF
\ENDFOR
\STATE Return
\[
\theta^*
=
\arg\max_{\theta \in \Theta}
Acc(\theta,S^*)
\]
\end{algorithmic}

\end{tcolorbox}
\endgroup

\subsection{Predictive Models}

To demonstrate the generality of the proposed framework, experiments are conducted across tabular, text, and image classification tasks. Decision Tree classifiers are employed for tabular datasets, TF-IDF features with a PyTorch-based multilayer perceptron (MLP) are used for text datasets, and ResNet18 is employed for image classification datasets. This combination enables evaluation across conventional machine learning, neural text classification, and deep image classification settings.

\subsection{Sensitivity to the Number of Anchor Configurations}
\label{sec:anchor_sensitivity}
To assess dependence on the number of construction anchors, we evaluated $m\in\{10,20,30\}$. Within each dataset--seed
combination, the anchor sets were nested: the 10-anchor set was contained in the 20-anchor set, which was contained in the 30-anchor set. A separate RPC coreset was constructed
for each anchor count, including $m=20$. Each resulting coreset was evaluated on the same fixed set of 20 held-out configurations, disjoint from all construction anchors,
using the same full-data reference scores and validation partition. Thus, anchor count varied while the held-out evaluation pool remained unchanged.

We used the same seeds as in the main experiments. Results were averaged across seeds within each dataset, with equal
weight given to each dataset in the modality-level summaries. Construction times are excluded from this sensitivity analysis
because differences in cache reuse and run resumption make the recorded elapsed times unsuitable for a direct comparison across anchor counts.

Table~\ref{tab:anchor_sensitivity_tradeoff} summarizes the results by modality. Larger anchor sets produced higher average held-out ranking correlations for tabular and text
datasets, suggesting that additional anchors may provide broader coverage of configuration behaviour. However, the response varied across datasets. For both image datasets,
20 anchors yielded the highest mean held-out correlation. The image-level mean retained fraction was also lowest at 20 anchors. These findings support retaining 20 anchors as
a practical intermediate setting, rather than a universally optimal choice.

\begin{table}[htbp]
\centering
\small
\caption{Anchor-count sensitivity summarized by modality.}
\label{tab:anchor_sensitivity_tradeoff}
\begin{tabular}{llrrrr}
\hline
Modality & Anchors & Anchor $\rho$ & Held-out $\rho$
& Retained (\%) \\
\hline
Tabular & 10 & 0.921 & 0.705 & 20.07 \\
        & 20 & 0.894 & 0.729 & 20.74  \\
        & 30 & 0.871 & 0.750 & 21.40  \\
\hline
Text    & 10 & 0.648 & 0.305 & 25.20 \\
        & 20 & 0.644 & 0.357 & 27.38  \\
        & 30 & 0.616 & 0.387 & 27.59  \\
\hline
Image & 10 & 0.890 & 0.684 & 21.57 \\
      & 20 & 0.853 & 0.733 & 15.46 \\
      & 30 & 0.838 & 0.713 & 19.33 \\
\hline
\end{tabular}
\medskip
\begin{minipage}{\linewidth}
\footnotesize
\textit{Note:} Values are averaged across seeds within each dataset and then equally across datasets within each modality.
\end{minipage}
\end{table}

Table~\ref{tab:sensitivity_overall_tests} reports exploratory Wilcoxon comparisons using paired dataset-level mean held-out correlations across 11 datasets. The 10-versus-30 comparison yielded an unadjusted $p=0.0186$, whereas the 10-versus-20 and 20-versus-30 comparisons yielded $p=0.0674$ and
$p=0.4648$, respectively. These results should be interpreted cautiously because the three comparisons were not adjusted for multiple testing. Dataset-level results are provided in~\ref{app:anchor_sensitivity},
Table~\ref{tab:sensitivity_dataset_detail}.

\begin{table}[htbp]
\centering
\small
\caption{Overall comparisons of construction-anchor counts
using two-sided Wilcoxon signed-rank tests on paired
dataset-level mean RPC held-out Spearman correlations
($n=11$).}
\label{tab:sensitivity_overall_tests}
\begin{tabular}{lcc}
\hline
Anchor-count comparison & $p$-value \\
\hline
10 versus 20 & 0.0674 \\
20 versus 30 & 0.4648 \\
10 versus 30 & 0.0186 \\
\hline
\end{tabular}
\end{table}

\subsection{Sensitivity to the Size-Penalty Coefficient}
\label{sec:alpha_sensitivity}

We examined $\alpha \in \{0,0.10,0.20,0.40\}$ on Adult and
IMDb using seeds 42, 43, and 44. Within each dataset--seed
combination, the training and validation partitions, 20
construction anchors, 20 disjoint held-out configurations,
and full-data reference scores were fixed across alpha values.
A separate coreset was evolved for each value, including
$\alpha=0.20$, using the same search settings and initial
random seed. Held-out scores were used only for evaluation,
not for coreset selection. 
Detailed results are reported in
~\ref{app:alpha_sensitivity},
Table~\ref{tab:alpha_sensitivity}.

The alpha-sensitivity analysis shows that the size penalty influences the observed ranking–size trade-off. Removing the penalty yielded the highest mean held-out correlation on both datasets, but retained larger subsets. Positive penalties reduced the mean retained fraction relative to the unpenalized setting, although the response was not monotonic on Adult. The default \(\alpha=0.20\) was a fixed heuristic setting and was not consistently superior to the alternatives. These descriptive results, obtained from separately evolved coresets on two datasets with three seeds each, do not establish a universally optimal penalty value. The main experimental results retain the original fixed setting of \(\alpha=0.20\).

\subsection{Computational Efficiency Analysis}
The computational effort of HPO is influenced by both the number of
hyperparameter evaluations and the amount of training data processed during
each evaluation. The following computational-cost expressions are simplified
theoretical approximations. Actual runtime also depends on model
architecture, hardware, preprocessing, and implementation overhead.
Let $k$ denote the total number of hyperparameter evaluations performed during optimization. For a single full-dataset HPO run, the
computational cost can be approximated as

\begin{equation}
C_{\mathrm{full}} \propto kN,
\end{equation}

where $N$ is the size of the full training set. Using the selected coreset, the per-run optimization cost can similarly be approximated as

\begin{equation}
C_{\mathrm{core}} \propto k|S^*|,
\end{equation}

where $|S^*| \ll N$.

Let $T_{\mathrm{full}}$ denote the measured wall-clock runtime of an HPO run in seconds
using the full training set, $T_{\mathrm{core}}$ the corresponding runtime
using the selected coreset, and $T_{\mathrm{sample}}$ the one-time runtime
required to construct the coreset. Here, $T_{\mathrm{sample}}$ measures the evolutionary
coreset-search time. The reported cost comparisons assume that the full-data anchor reference scores are already
available and exclude their initial evaluation cost. The per-run HPO-stage speedup is therefore

\begin{equation}
\label{eq:speedup}
\mathrm{Speedup}
=
\frac{T_{\mathrm{full}}}{T_{\mathrm{core}}}.
\end{equation}

Since the coreset construction stage is performed only once and can be amortized over repeated HPO runs, the break-even point is defined as

\begin{equation}
\label{eq:breakeven}
N_{\mathrm{BE}}
=
\frac{T_{\mathrm{sample}}}
{T_{\mathrm{full}}-T_{\mathrm{core}}}.
\end{equation}

where $N_{\mathrm{BE}}$ denotes the number of optimization runs required for the cumulative savings obtained from coreset-based optimization to offset the initial coreset construction cost.

\subsection{Held-Out Configuration Generalization Analysis}
\label{sec:heldout_config}

The proposed coreset is selected by preserving the ranking of a fixed set of anchor hyperparameter configurations. To examine whether this ranking preservation generalizes beyond the configurations used during coreset construction, we perform a held-out configuration analysis.

Let $C_A=\{c_1,c_2,\ldots,c_m\}$ denote the anchor configuration set used during coreset selection, and let $C_H=\{h_1,h_2,\ldots,h_q\}$ denote a disjoint set of held-out configurations sampled from the same hyperparameter search space. In our experiments, $m=20$ anchor configurations and $q=20$ held-out configurations are used. For each experimental seed, a separate held-out configuration set
$C_H$ is generated such that $C_A \cap C_H = \emptyset$. After the coreset $S^*$ has been fixed, each held-out configuration
$h_j \in C_H$ is evaluated on the full training data and on the selected
coreset using the same validation set. The same held-out configurations
are also evaluated on the same-size random subset and, where applicable, the gradient-based subset under that seed.

The full-data held-out performance vector is defined as
\begin{equation}
P_{H} =
[f(h_1,D_{train},D_{val}),\ldots,f(h_q,D_{train},D_{val})],
\end{equation}
and the coreset-based held-out performance vector is defined as
\begin{equation}
P_{S^*} =
[f(h_1,S^*,D_{val}),\ldots,f(h_q,S^*,D_{val})].
\end{equation}

The held-out ranking correlation is then computed as
\begin{equation}
\rho_H =
\rho\left(
\operatorname{rank}(P_H),
\operatorname{rank}(P_{S^*})
\right).
\end{equation}

This analysis tests whether the ranking-preservation behavior of the selected coreset generalizes beyond the anchor configurations used during coreset construction. For comparison, the same held-out configurations are also evaluated on a random subset of the same size and, where applicable, on a fixed-budget gradient-based subset, allowing their held-out ranking correlations with the full-data ranking to be assessed under the same experimental conditions.

\section{Experiments}
\label{sec:experiments}

This section presents the experimental evaluation of the proposed ranking-preserving coreset selection framework. The experiments aim to assess whether the selected coresets can preserve the hyperparameter configuration rankings induced by the full dataset, while also reducing computational cost and maintaining competitive predictive performance across diverse benchmark datasets.

\subsection{Datasets}
We evaluate the proposed approach on 11 benchmark datasets, including seven tabular datasets from the UCI repository, two text classification datasets, and two image classification datasets.

The tabular datasets include
Dermatology\footnote{\url{https://doi.org/10.24432/C5FK5P}},
Car Evaluation\footnote{\url{https://doi.org/10.24432/C5JP48}},
Abalone\footnote{\url{https://doi.org/10.24432/C55C7W}},
Adult\footnote{\url{https://doi.org/10.24432/C5XW20}},
Diabetes 130-US Hospitals\footnote{\url{https://doi.org/10.24432/C5230J}},
Census-Income\footnote{\url{https://doi.org/10.24432/C5GP7S}},
and CoverType\footnote{\url{https://doi.org/10.24432/C50K5N}}.
These datasets cover a wide range of characteristics, including varying numbers of instances, feature dimensions, class distributions, and classification difficulty.

For text sentiment classification, we use two benchmark datasets:
IMDb\footnote{\url{https://huggingface.co/datasets/stanfordnlp/imdb}}
and TweetEval Sentiment(TweetEval)\footnote{
\url{https://huggingface.co/datasets/cardiffnlp/tweet_eval}},
representing binary and multi-class sentiment classification tasks, respectively. These datasets capture real-world linguistic variability in user-generated text. Each document was represented using TF-IDF features and classified using a PyTorch-based multilayer perceptron (MLP), enabling efficient evaluation of the proposed ranking-preserving coreset framework on high-dimensional text data.

For image classification, we use
CIFAR-10\footnote{\url{https://www.cs.toronto.edu/~kriz/cifar.html}}
and DermaMNIST\footnote{\url{https://medmnist.com/}}. CIFAR-10 is a widely used natural image classification benchmark consisting of 10 object categories, while DermaMNIST is a medical image classification dataset derived from dermatoscopic skin lesion images and contains 7 diagnostic classes. These datasets enable evaluation of the proposed framework in both general-purpose computer vision and medical imaging settings. 

Table~\ref{tab:datasets} provides a summary of the number of instances, features, and classes of all datasets used in our experiments. Class counts and percentages are provided in~\ref{app:class_distributions}.

\subsection{Data Preprocessing}
Missing-value markers were converted to missing entries.
Numerical features were median-imputed and standardized;
categorical features were imputed using the most frequent value
and one-hot encoded, with unknown categories ignored.
The preprocessor was fitted on the training partition and applied
to the validation and test partitions. Targets were label-encoded.
When stratified splitting failed because of rare classes, labels
occurring fewer than twice in the partition being split were
merged into its most frequent class before retrying the split.

For each text dataset, available official partitions were combined,
records with missing text or labels were removed, and a stratified
sample of 10,000 documents was selected before the experimental
60\%/20\%/20\% split. TF-IDF used lowercasing, Unicode accent
stripping, and default word tokenization, without explicit
stop-word removal, stemming, or lemmatization. Vocabulary size,
n-gram range, document-frequency thresholds, and sublinear term
frequency were tuned. Each vectorizer was fitted on the training
data used for that evaluation and applied to validation or test data.

CIFAR-10 images were represented as $32 \times 32 \times 3$ inputs, converted to tensors, and normalized using
channel means $(0.4914,0.4822,0.4465)$ and standard deviations
$(0.2470,0.2435,0.2616)$. DermaMNIST images were loaded as RGB,
resized to $64 \times 64$, converted to tensors, and normalized
using a mean and standard deviation of $0.5$ per channel.
No random cropping or flipping was applied.

Stratified sampling with fixed random seeds is used to partition datasets. For the tabular and text datasets, 60\% of the data was used for training, 20\% for validation, and 20\% for testing. The same validation partition was used during both coreset construction and hyperparameter optimization, while the test partition remained unseen until the final evaluation. For the image datasets, the official test partitions were retained for final testing, and the available training data were split into training and validation subsets (75\%/25\%). For CIFAR-10, this split used the official training partition; for DermaMNIST, it used the combined
official training and validation partitions.

\subsection{Experimental Settings}
To ensure robust evaluation, the main experiments were repeated using multiple random seeds. The results were averaged over 10 independent runs for small and medium-sized tabular datasets, 5 independent runs for larger tabular and text datasets, and 3 independent runs for image datasets due to the higher computational cost of training ResNet18. The larger tabular datasets include Diabetes, Census-Income, and CoverType.

We evaluate the following settings for each dataset:

\begin{itemize}
    \item \textbf{Baseline:} Model training performed using the default hyperparameter configuration without hyperparameter optimization.

    \item \textbf{Full Training Data:} Hyperparameter optimization was performed on the complete training dataset. This setting serves as the full-data HPO reference.

    \item \textbf{Coreset (Proposed):} Hyperparameter optimization was performed on the evolved ranking-preserving coreset selected by the proposed coreset selection framework.

    \item \textbf{Random Subset:} Hyperparameter optimization performed on a stratified random subset with the same size and class-wise proportions as the proposed coreset. 

    \item \textbf{Gradient-Based Coreset:} A fixed-budget gradient-based subset is selected following the GRAD-MATCH principle~\cite{21}, which is also used within the AutoMata framework for compute-efficient HPO~\cite{5}. After a short warm start, final-layer gradients are computed, and positive OMP is applied class-wise to select a weighted subset. The subset budget and class-wise proportions are matched to the proposed RPC coreset, and the same selected subset is reused across GA, CHC, and PSO.
    
\end{itemize}

The proposed framework does not enforce a fixed coreset size. Instead, the subset size is implicitly controlled by the evolutionary process via the fitness function (Eq.~\ref{eq:fitness}), which rewards preserving rankings while penalizing larger subsets. Consequently, the final coreset size emerges naturally from the evolutionary search process and may vary across datasets and runs.

For ranking preservation analysis, the proposed coreset is compared with same-size subset baselines. A stratified random subset is generated with the same size and class-wise proportions as the evolved coreset. In addition, for differentiable text and image classification models, a gradient-preserving subset is selected using the same budget. The resulting rankings are evaluated using Spearman's rank correlation, the agreement ratio, and MAE. The gradient-based baseline is not applied to the Decision Tree tabular experiments because Decision Trees are non-differentiable and do not provide the gradient information required for gradient matching.

For all datasets, identical training, validation, and test partitions were used throughout all experiments to ensure a fair comparison. The proposed coreset and random reference subset were both constructed from the same training partition, while the validation and test sets remained unchanged across all experimental settings.

All results are reported as the mean and standard deviation to capture stability across runs.

\begin{table}[h]
\centering
\footnotesize
\caption{Summary of datasets used in the experiments.}
\label{tab:datasets}
\begin{tabularx}{\textwidth}{Xcccc}
\hline
\textbf{Dataset} & \textbf{Type} & \textbf{Instances Used} & \textbf{Features} & \textbf{Classes} \\
\hline
Dermatology & Tabular & 366 & 34 & 6 \\
Car Evaluation & Tabular & 1,728 & 6 & 4 \\
Abalone & Tabular & 4,177 & 8 & 28 \\
Adult & Tabular & 32,561 & 14 & 2 \\
Diabetes & Tabular & 101,766 & 49 & 3 \\
Census-Income & Tabular & 199,523 & 41 & 2 \\
CoverType & Tabular & 581,012 & 54 & 7 \\
\hline
IMDb & Text & 10,000 & Raw text & 2 \\
TweetEval  & Text & 10,000 & Raw text & 3 \\
\hline
CIFAR-10 & Image & 60,000 & $32 \times 32 \times 3$ & 10 \\
DermaMNIST & Image & 10,015 & $64 \times 64 \times 3$ & 7 \\
\hline
\end{tabularx}
\end{table}

\subsection{Implementation and Computing Environment}
\label{sec:environment}
Experiments were conducted on a workstation running Windows 11 Pro (64-bit), equipped with a 13th-generation Intel Core i7-13700 CPU, 32 GB RAM, and an NVIDIA RTX A4000 GPU with 16 GB VRAM.
The implementation used Python 3.12.13 in an Anaconda environment with Jupyter Notebook 7.5.5. The software environment comprised
PyTorch 2.5.1 with CUDA 12.1, torchvision 0.20.1, scikit-learn 1.8.0, NumPy 2.3.1, pandas 3.0.2, Hugging Face Datasets 4.8.5, and MedMNIST 3.0.2.

\subsection{Hyperparameter Optimization Algorithms}

To evaluate the robustness of the proposed framework, hyperparameter optimization was performed using three evolutionary optimization algorithms: Genetic Algorithm (GA), Cross-generational elitist
selection, Heterogeneous recombination, and Cataclysmic mutation (CHC),
and Particle Swarm Optimization (PSO).

For each optimizer, identical search budgets were used across the Full Training Data, Proposed Coreset, Random subset, and Gradient-based subset to ensure a fair comparison. After optimization, the best hyperparameter configuration, as determined by validation performance, is used to train the final model, which is then evaluated on the held-out test set.

\subsection{Experimental Protocol and Search Space}

For each dataset, hyperparameter optimization was performed using the same search space and evaluation budget across Full HPO and the proposed Ranking-Preserving Coreset HPO. This ensures that performance differences are attributable to the training data used during optimization rather than to differences in search budget. The validation set was used to evaluate candidate hyperparameter configurations during both ranking-preserving coreset selection and downstream HPO, while the test set was used only for final performance evaluation.

The baseline setting involves training each model with a fixed default hyperparameter configuration without performing HPO. This baseline is included to quantify the performance gain obtained from hyperparameter optimization and to assess whether coreset-based HPO can retain the benefits of full-dataset HPO.

The hyperparameter search spaces and optimization budgets used in the experiments are summarized in Tables~\ref{tab:search_space} and~\ref{tab:optimization_budget}, respectively.

\begin{table}[htbp]
\centering

\caption{Hyperparameter search spaces used for HPO across tabular, text, and image datasets.}
\label{tab:search_space}
\begin{tabularx}{\textwidth}{
    @{}p{1.3cm}
    p{2.2cm}
    >{\raggedright\arraybackslash}X@{}
}
\hline
\textbf{Data Type} & \textbf{Model} & \textbf{Search Space} \\
\hline

Tabular 
& Decision Tree 
& \begin{tabular}[t]{@{}p{\linewidth}@{}}
criterion $\in$ \{gini, entropy, log\_loss\} \\
max\_depth $\in$ \{None, 5, 10, 15, 20, 30\} \\
min\_samples\_split $\in$ \{2, 5, 10, 20\} \\
min\_samples\_leaf $\in$ \{1, 2, 5, 10\} \\
max\_features $\in$ \{None, sqrt, log2\}
\end{tabular} \\

\hline

Text
& TF-IDF + PyTorch MLP 
& \begin{tabular}[t]{@{}p{\linewidth}@{}}
max\_features $\in$ \{5000, 10000, 20000\} \\
ngram\_range $\in$ \{(1,1), (1,2)\} \\
min\_df $\in$ \{1, 2, 5\} \\
max\_df $\in$ \{0.90, 0.95, 1.0\} \\
sublinear\_tf $\in$ \{True, False\} \\
hidden\_layers $\in$ \{(64), (128), (256), (512), (128,64), (256,128), (512,256)\} \\
dropout $\in$ \{0.0, 0.2, 0.4\} \\
activation $\in$ \{relu, tanh, gelu\} \\
learning\_rate $\in$ \{$10^{-4}$, $3\times10^{-4}$, $10^{-3}$, $3\times10^{-3}$\} \\
weight\_decay $\in$ \{0, $10^{-5}$, $10^{-4}$, $10^{-3}$\} \\
batch\_size $\in$ \{64, 128\} \\
epochs $\in$ \{5, 10, 20\} \\
optimizer $\in$ \{Adam, AdamW\}
\end{tabular} \\

\hline

Image 
& ResNet18 
& \begin{tabular}[t]{@{}p{\linewidth}@{}}
learning\_rate $\in$ \{$10^{-5}$, $10^{-4}$, $3\times10^{-4}$, $10^{-3}$, $3\times10^{-3}$\} \\
weight\_decay $\in$ \{0, $10^{-5}$, $10^{-4}$, $5\times10^{-4}$\} \\
batch\_size $\in$ \{64, 128, 256\} \\
optimizer $\in$ \{SGD, AdamW\} \\
epochs $\in$ \{3, 5, 8, 12\}
\end{tabular} \\

\hline
\end{tabularx}
\end{table}

\begin{table}[htbp]
\centering
\caption{Coreset selection and HPO budgets used in the experiments.}
\label{tab:optimization_budget}
\begin{tabular}{lccc}
\hline
\textbf{Parameter} & \textbf{Tabular} & \textbf{Text} & \textbf{Image} \\
\hline
Number of anchor configurations $m$ & 20 & 20 & 20 \\
Anchor candidate pool size $M$ & 500 & 500 & 500 \\
Elite count $E$ & 1 & 1 & 1 \\
Crossover probability $p_c$ & 1.0 & 1.0 & 1.0 \\
Coreset mutation rate $p_m$ & 0.10 & 0.10 & 0.10 \\
Minimum coreset fraction $f_{\min}$ & 0.10 & 0.10 & 0.10 \\
Maximum coreset fraction $f_{\max}$ & 0.30 & 0.30 & 0.30 \\
Size penalty coefficient $\alpha$ & 0.20 & 0.20 & 0.20 \\
Coreset GA population size & 10 & 10 & 6 \\
Coreset GA maximum generations & 20 & 20 & 12 \\
Coreset GA patience & 5 & 5 & 4 \\
HPO population size & 20 & 15 & 8 \\
HPO maximum generations & 20 & 30 & 12 \\
HPO patience & 5 & 8 & 4 \\
GA mutation rate & 0.30 & 0.30 & 0.30 \\
CHC cataclysmic mutation rate & 0.35 & 0.35 & 0.35 \\
PSO inertia weight & 0.7298 & 0.7298 & 0.7298 \\
PSO cognitive coefficient & 1.49618 & 1.49618 & 1.49618 \\
PSO social coefficient & 1.49618 & 1.49618 & 1.49618 \\
\hline
\end{tabular}
\end{table}

For the Random Subset baseline, the subset size and class-wise proportions are matched to the evolved coreset in each run. For the gradient-based baseline, the same subset budget is used for differentiable text and image classification models; it is not applied to the Decision Tree tabular experiments because Decision Trees are non-differentiable and do not provide gradients required for gradient matching.

\subsection{Evaluation Metrics}

The proposed framework is evaluated using three types of metrics: final classification performance, ranking preservation, and computational efficiency. The main evaluation metrics are:
\begin{itemize}[noitemsep]
    \item Test accuracy of the final model,
    \item Spearman rank correlation between the full-data ranking and each subset-induced ranking,
    \item Pairwise agreement ratio between the full-data ranking and each subset-induced ranking,
    \item Mean absolute error (MAE) between full-data and subset-based validation performance scores,
    \item Computational time required for HPO and coreset construction.
\end{itemize}

For ranking evaluation, a fixed set of anchor hyperparameter configurations was first evaluated on the full training dataset to obtain the reference full-data ranking. The same configurations were then evaluated on the proposed ranking-preserving coreset, the stratified random subset and, where applicable, the gradient-based subset. For each subset, Spearman rank correlation is used as the primary ranking-preservation metric for both the anchor and held-out configuration analyses. Pairwise agreement
and MAE are additionally reported as supplementary diagnostics for the anchor
configurations. The stratified random and gradient-based subsets were selected using the same subset size and class-wise proportions as the proposed coreset to ensure a fair comparison.

\subsection{Statistical Analysis}

To assess whether the proposed ranking-preserving coreset provides significantly different ranking preservation compared with baseline subset-selection methods, Wilcoxon signed-rank tests were performed on the Spearman rank correlation values obtained across datasets. For each comparison, correlations were first averaged across
seeds within each dataset. The Wilcoxon test then used paired dataset-level means, so the statistical unit was the dataset.

For each baseline subset method, the null and alternative hypotheses are defined as:

\begin{equation}
H_0:\;
\rho_{\text{proposed}}
=
\rho_{baseline}
\end{equation}

\begin{equation}
H_1:\;
\rho_{\text{proposed}}
\neq
\rho_{baseline}
\end{equation}

where $\rho_{\text{proposed}}$ denotes the Spearman rank correlation obtained using the proposed ranking-preserving coreset, and $\rho_{baseline}$ denotes the Spearman rank correlation obtained using a baseline subset method, i.e., a random subset and the gradient-based subset, where applicable.

The Wilcoxon signed-rank test was applied separately to assess anchor-based ranking preservation and generalization across held-out configurations. The anchor-based analysis evaluates whether the proposed coreset preserves the ranking of the configurations used during coreset selection, while the held-out analysis evaluates whether this preservation extends to non-anchor configurations not used during subset construction.

Statistical significance was assessed at the $\alpha=0.05$ level. Since the gradient-based coreset requires differentiable models, comparisons involving this baseline were conducted only for the text and image experiments.

\section{Results and Discussion}
\label{sec:results_discussion}

This section presents a comprehensive evaluation of the proposed ranking-preserving coreset selection framework. The results are analyzed across multiple dimensions, including hyperparameter optimization performance, ranking preservation quality, and computational efficiency. Experiments are conducted on seven UCI tabular datasets, two text classification datasets, and two image classification datasets, with results reported as mean $\pm$ standard deviation over multiple runs.

\subsection{Hyperparameter Optimization Performance}
\label{sec:hpo_performance}
Table~\ref{tab:baseline_fullhpo} first examines whether hyperparameter optimization provides an improvement over the default model configuration. Full-data HPO achieves higher mean test accuracy than the default configuration across all evaluated datasets, although the magnitude of improvement varies considerably. The largest gains are observed for Abalone, Adult, Diabetes, and CIFAR-10, where the best Full-HPO accuracy increases from $0.200$ to $0.262$, $0.814$ to $0.857$, $0.511$ to $0.587$, and $0.743$ to $0.817$, respectively. Improvements are also observed for Census-Income, IMDb, TweetEval, and DermaMNIST, while the gains for Dermatology, Car Evaluation, and CoverType are comparatively small. Overall, these results confirm that hyperparameter optimization is beneficial for most of the evaluated datasets, thereby motivating further investigation into whether comparable HPO outcomes can be achieved with substantially smaller training data subsets.

\begin{table}[htbp]
\centering
\footnotesize
\caption{Test-set classification accuracy of the default baseline and full-dataset hyperparameter optimization using GA, CHC, and PSO. Results are reported as mean $\pm$ standard deviation across seeds.}
\label{tab:baseline_fullhpo}
\setlength{\tabcolsep}{4pt}
\renewcommand{\arraystretch}{1.15}
\resizebox{\linewidth}{!}{%
\begin{tabular}{lcccc}
\hline
\noalign{\vskip 4pt}
\textbf{Dataset} &
\textbf{Baseline} &
\shortstack{\textbf{Full-HPO}\\\textbf{(GA)}} &
\shortstack{\textbf{Full-HPO}\\\textbf{(CHC)}} &
\shortstack{\textbf{Full-HPO}\\\textbf{(PSO)}} \\
\hline

Dermatology
& $0.947 \pm 0.030$
& $\boldsymbol{0.954 \pm 0.031}$
& $0.938 \pm 0.027$
& $\boldsymbol{0.954 \pm 0.028}$ \\

Car Evaluation
& $0.955 \pm 0.009$
& $0.962 \pm 0.011$
& $\boldsymbol{0.964 \pm 0.007}$
& $0.958 \pm 0.017$ \\

Abalone
& $0.200 \pm 0.016$
& $\boldsymbol{0.262 \pm 0.011}$
& $0.259 \pm 0.010$
& $0.258 \pm 0.021$ \\

Adult
& $0.814 \pm 0.003$
& $\boldsymbol{0.857 \pm 0.005}$
& $0.855 \pm 0.005$
& $\boldsymbol{0.857 \pm 0.005}$ \\

Diabetic
& $0.511 \pm 0.002$
& $\boldsymbol{0.587 \pm 0.002}$
& $0.585 \pm 0.004$
& $0.585 \pm 0.004$ \\

Census-Income
& $0.930 \pm 0.002$
& $\boldsymbol{0.950 \pm 0.000}$
& $\boldsymbol{0.950 \pm 0.001}$
& $\boldsymbol{0.950 \pm 0.000}$ \\

CoverType
& $0.930 \pm 0.001$
& $0.934 \pm 0.001$
& $\boldsymbol{0.935 \pm 0.001}$
& $0.933 \pm 0.005$ \\

\hline

IMDb
& $0.872 \pm 0.005$
& $\boldsymbol{0.893 \pm 0.007}$
& $\boldsymbol{0.893 \pm 0.006}$
& $\boldsymbol{0.893 \pm 0.007}$ \\

TweetEval
& $0.578 \pm 0.012$
& $0.598 \pm 0.010$
& $\boldsymbol{0.600 \pm 0.012}$
& $0.594 \pm 0.010$ \\

\hline

CIFAR-10
& $0.743 \pm 0.008$
& $0.811 \pm 0.008$
& $0.816 \pm 0.004$
& $\boldsymbol{0.817 \pm 0.002}$ \\

DermaMNIST
& $0.722 \pm 0.008$
& $0.741 \pm 0.005$
& $\boldsymbol{0.756 \pm 0.011}$
& $0.751 \pm 0.006$ \\
\hline
\end{tabular}%
}
\end{table}
\begin{figure}[htbp]
    \centering
    \includegraphics[width=\textwidth]{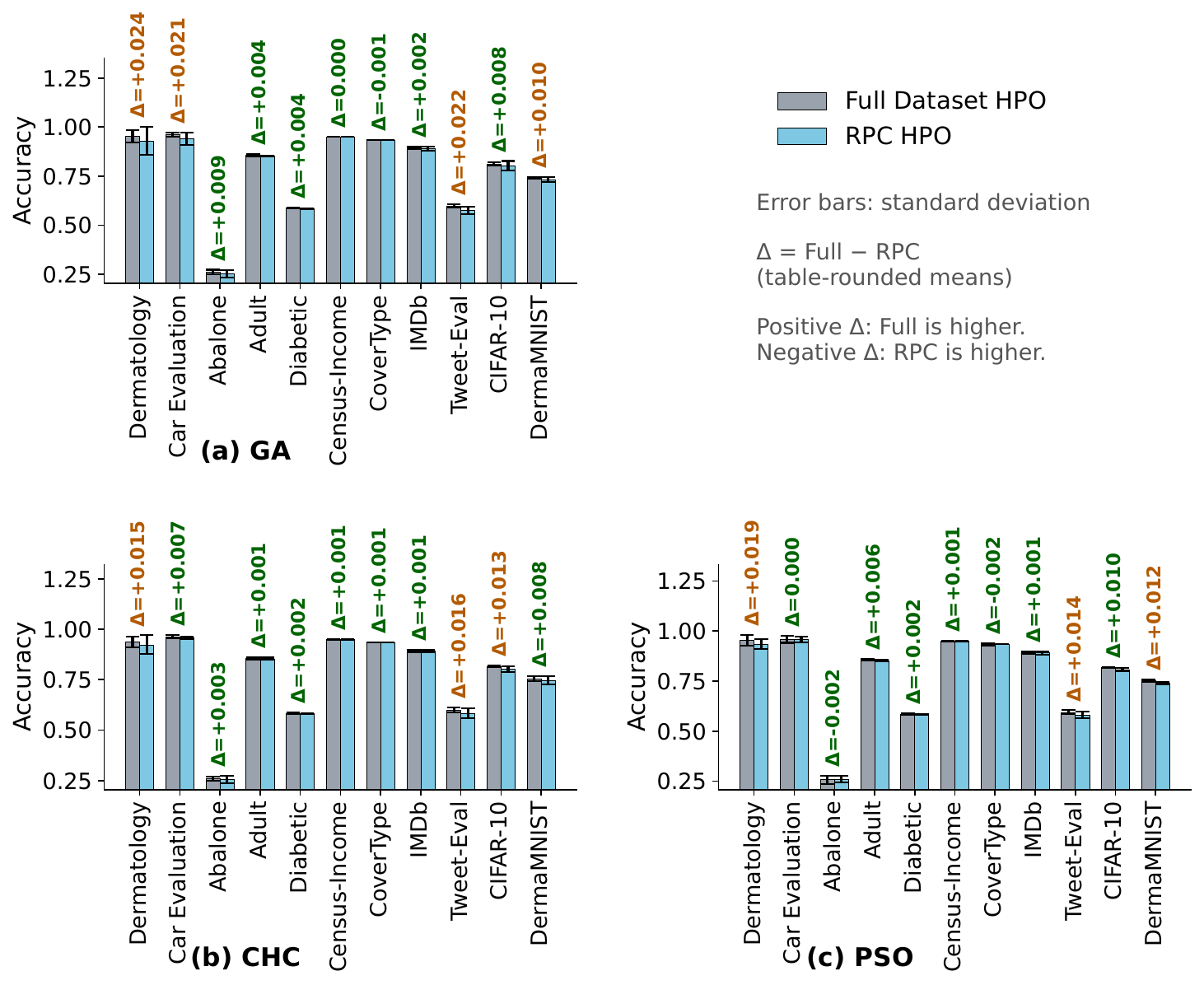}
    \caption{Comparison of test-set classification accuracy obtained using full-dataset HPO and ranking-preserving coreset (RPC) HPO across different optimizers. Error bars indicate standard deviation across multiple runs. The annotated values ($\Delta$) represent the accuracy difference between Full HPO and RPC HPO.}
    \label{fig:accuracy}
\end{figure}

\begin{figure}[h]
    \centering
    \includegraphics[width=\textwidth]{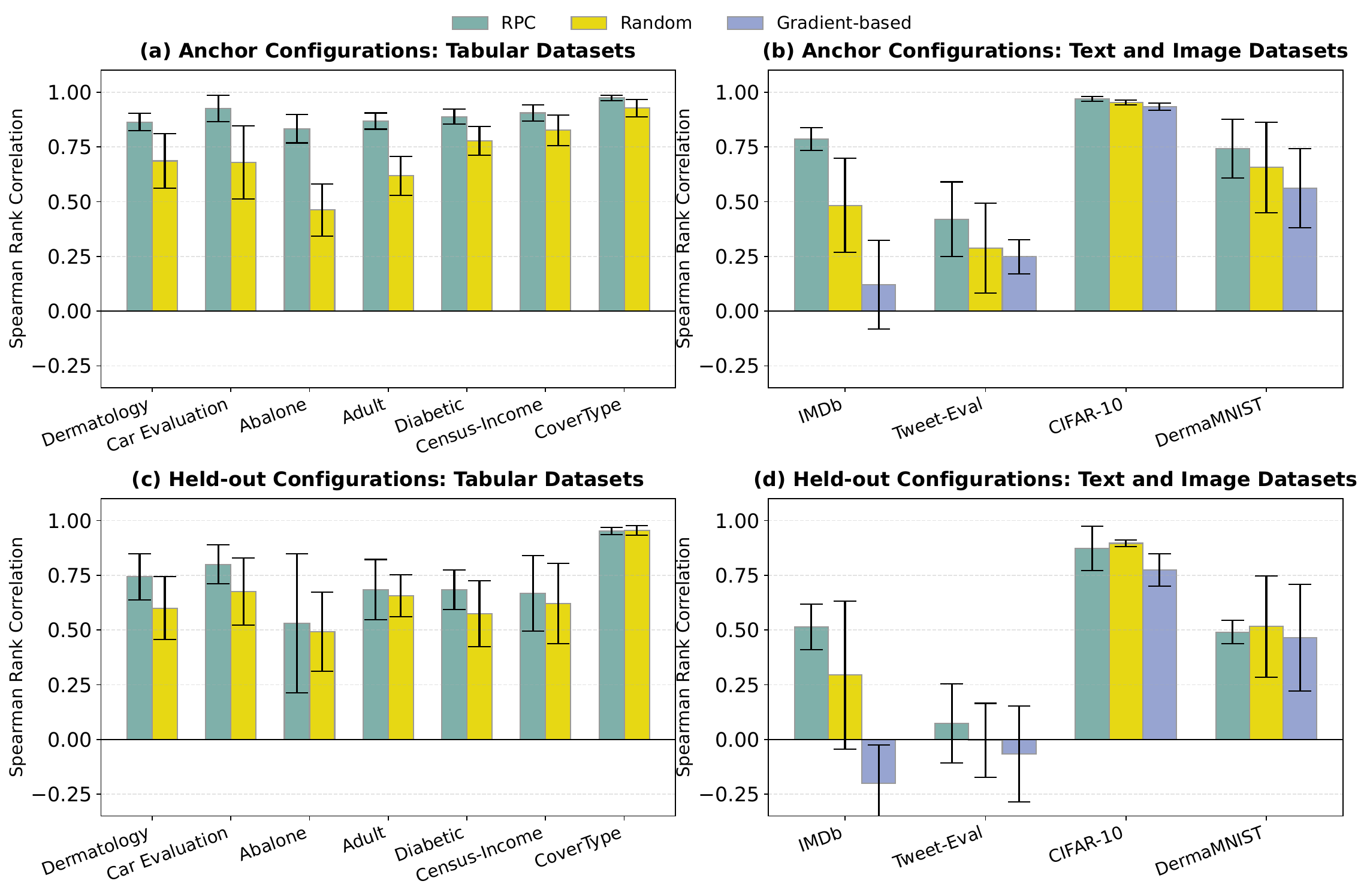}
    \caption{Spearman rank correlation between full-data and subset-induced configuration rankings. Panels (a) and (b) show in-sample fit on the construction anchors for tabular and text/image datasets, respectively; panels (c) and (d) show ranking preservation on held-out configurations. Error bars show standard deviation across seeds.}
    \label{fig:spearman}
\end{figure}

Tables~\ref{tab:ga_accuracy}--\ref{tab:pso_accuracy} present the detailed test-accuracy results across the three optimizers, including comparisons with the random and gradient-based baselines. Full HPO represents optimization on the complete training dataset, whereas RPC HPO uses the proposed ranking-preserving coreset. For comparison, Random HPO uses a same-size stratified random subset, while Gradient-Based HPO uses a same-size gradient-based coreset and is reported only for datasets to which this baseline is applicable. Figure~\ref{fig:accuracy} complements these results by visually illustrating the close agreement in predictive performance between Full HPO and RPC HPO. Results are reported as mean $\pm$ standard deviation across seeds. Overall, RPC HPO maintains test accuracy close to Full HPO while using substantially fewer training instances. The mean absolute difference between Full HPO and RPC HPO is approximately $0.73$ percentage points; 30 of the 33 differences are within two percentage points, and the largest observed difference is approximately $2.4$ percentage points. A direct comparison with the same-sized stratified random baseline provides a more nuanced view of predictive performance. Across 33 dataset–optimizer combinations, RPC HPO achieves higher mean test accuracy in 19 cases, lower in 11, and the same in 3. The differences are generally small relative to the reported run-to-run variability, and no consistent accuracy advantage of RPC over random sampling is evident. It is important to note, however, that the random baseline does not determine its subset size independently. For each run, its sampling budget and class-wise proportions are matched to those produced by RPC; therefore, random sampling serves as a matched-budget control for evaluating the effect of subset composition. Once this RPC-derived budget is specified, constructing the random subset is computationally inexpensive. Thus, the contribution of RPC is not an improvement in final predictive accuracy over random sampling at a matched budget, but its ability to construct a compact subset that more faithfully preserves the full-data HPO landscape. As shown in Tables~\ref{tab:ranking} and~\ref{tab:heldout_ranking} and visually summarized in Figure~\ref{fig:spearman}, RPC generally achieves stronger Spearman rank correlations than same-size random sampling, indicating more reliable preservation of the relative ordering of hyperparameter configurations. RPC also has higher mean Spearman correlation than the gradient-based baseline on all four applicable datasets. For held-out configurations, however, random sampling achieves slightly higher correlations for image datasets, indicating that the advantage of RPC does not extend uniformly to every dataset. 


\begin{table}[htbp]
\centering
\footnotesize
\caption{Test-set classification accuracy obtained using GA-based hyperparameter optimization on the full training data (Full HPO), the proposed ranking-preserving coreset (RPC HPO), a same-size stratified random subset (Random HPO), and a same-size fixed-budget gradient-based coreset (Gradient-Based HPO). Results are reported as mean $\pm$ standard deviation across seeds. Gradient-Based HPO applies only to the differentiable text and image classification models.}
\label{tab:ga_accuracy}

\setlength{\tabcolsep}{5pt}
\renewcommand{\arraystretch}{1.15}

\resizebox{\linewidth}{!}{%
\begin{tabular}{lcccc}
\hline
\noalign{\vskip 2pt}
\textbf{Dataset} &
\shortstack{\textbf{Full}\\\textbf{HPO}} &
\shortstack{\textbf{RPC}\\\textbf{HPO}} &
\shortstack{\textbf{Random}\\\textbf{HPO}} &
\shortstack{\textbf{Gradient-Based}\\\textbf{HPO}} \\
\noalign{\vskip 2pt}
\hline

Dermatology
& $0.954 \pm 0.031$
& $0.930 \pm 0.071$
& $0.939 \pm 0.036$
& -- \\

Car Evaluation
& $0.962 \pm 0.011$
& $0.941 \pm 0.031$
& $0.942 \pm 0.027$
& -- \\

Abalone
& $0.262 \pm 0.011$
& $0.253 \pm 0.019$
& $0.247 \pm 0.026$
& -- \\

Adult
& $0.857 \pm 0.005$
& $0.853 \pm 0.003$
& $0.851 \pm 0.008$
& -- \\

Diabetes
& $0.587 \pm 0.002$
& $0.583 \pm 0.001$
& $0.583 \pm 0.001$
& -- \\

Census-Income
& $0.950 \pm 0.000$
& $0.950 \pm 0.001$
& $0.949 \pm 0.002$
& -- \\

CoverType
& $0.934 \pm 0.001$
& $0.935 \pm 0.001$
& $0.929 \pm 0.013$
& -- \\

\hline
IMDb
& $0.893 \pm 0.007$
& $0.891 \pm 0.010$
& $0.886 \pm 0.011$
& $0.878 \pm 0.017$ \\

TweetEval
& $0.598 \pm 0.010$
& $0.576 \pm 0.020$
& $0.591 \pm 0.015$
& $0.584 \pm 0.013$ \\

\hline
CIFAR-10
& $0.811 \pm 0.008$
& $0.803 \pm 0.025$
& $0.786 \pm 0.021$
& $0.797 \pm 0.017$ \\

DermaMNIST
& $0.741 \pm 0.005$
& $0.731 \pm 0.012$
& $0.734 \pm 0.012$
& $0.736 \pm 0.001$ \\

\hline
\end{tabular}%
}
\end{table}

\begin{table}[h]
\centering
\footnotesize
\caption{Test-set classification accuracy obtained using CHC-based hyperparameter optimization on the full training data (Full HPO), the proposed ranking-preserving coreset (RPC HPO), a same-size stratified random subset (Random HPO), and a same-size fixed-budget gradient-based coreset (Gradient-Based HPO). Results are reported as mean $\pm$ standard deviation across seeds. Gradient-Based HPO applies only to the differentiable text and image classification models.}
\label{tab:chc_accuracy}

\setlength{\tabcolsep}{5pt}
\renewcommand{\arraystretch}{1.15}

\resizebox{\linewidth}{!}{%
\begin{tabular}{lcccc}
\hline
\noalign{\vskip 2pt}
\textbf{Dataset} &
\shortstack{\textbf{Full}\\\textbf{HPO}} &
\shortstack{\textbf{RPC}\\\textbf{HPO}} &
\shortstack{\textbf{Random}\\\textbf{HPO}} &
\shortstack{\textbf{Gradient-Based}\\\textbf{HPO}} \\
\noalign{\vskip 2pt}
\hline

Dermatology
& $0.938 \pm 0.027$
& $0.923 \pm 0.047$
& $0.931 \pm 0.045$
& -- \\

Car Evaluation
& $0.964 \pm 0.007$
& $0.957 \pm 0.007$
& $0.943 \pm 0.026$
& -- \\

Abalone
& $0.259 \pm 0.010$
& $0.256 \pm 0.017$
& $0.244 \pm 0.025$
& -- \\

Adult
& $0.855 \pm 0.005$
& $0.854 \pm 0.005$
& $0.851 \pm 0.007$
& -- \\

Diabetes
& $0.585 \pm 0.004$
& $0.583 \pm 0.002$
& $0.583 \pm 0.001$
& -- \\

Census-Income
& $0.950 \pm 0.001$
& $0.949 \pm 0.001$
& $0.950 \pm 0.001$
& -- \\

CoverType
& $0.935 \pm 0.001$
& $0.934 \pm 0.001$
& $0.935 \pm 0.001$
& -- \\

\hline
IMDb
& $0.893 \pm 0.006$
& $0.892 \pm 0.007$
& $0.889 \pm 0.009$
& $0.873 \pm 0.015$ \\

TweetEval
& $0.600 \pm 0.012$
& $0.584 \pm 0.024$
& $0.586 \pm 0.017$
& $0.579 \pm 0.013$ \\

\hline
CIFAR-10
& $0.816 \pm 0.004$
& $0.803 \pm 0.013$
& $0.802 \pm 0.010$
& $0.762 \pm 0.052$ \\

DermaMNIST
& $0.756 \pm 0.011$
& $0.748 \pm 0.021$
& $0.755 \pm 0.006$
& $0.733 \pm 0.004$ \\

\hline
\end{tabular}%
}
\end{table}

\begin{table}[h]
\centering
\footnotesize
\caption{Test-set classification accuracy obtained using PSO-based hyperparameter optimization on the full training data (Full HPO), the proposed ranking-preserving coreset (RPC HPO), a same-size stratified random subset (Random HPO), and a same-size fixed-budget gradient-based coreset (Gradient-Based HPO). Results are reported as mean $\pm$ standard deviation across seeds. Gradient-Based HPO applies only to the differentiable text and image classification models.}

\label{tab:pso_accuracy}

\setlength{\tabcolsep}{5pt}
\renewcommand{\arraystretch}{1.15}

\resizebox{\linewidth}{!}{%
\begin{tabular}{lcccc}
\hline
\noalign{\vskip 2pt}
\textbf{Dataset} &
\shortstack{\textbf{Full}\\\textbf{HPO}} &
\shortstack{\textbf{RPC}\\\textbf{HPO}} &
\shortstack{\textbf{Random}\\\textbf{HPO}} &
\shortstack{\textbf{Gradient-Based}\\\textbf{HPO}} \\
\noalign{\vskip 2pt}
\hline

Dermatology
& $0.954 \pm 0.028$
& $0.935 \pm 0.024$
& $0.942 \pm 0.046$
& -- \\

Car Evaluation
& $0.958 \pm 0.017$
& $0.958 \pm 0.015$
& $0.948 \pm 0.020$
& -- \\

Abalone
& $0.258 \pm 0.021$
& $0.260 \pm 0.016$
& $0.249 \pm 0.022$
& -- \\

Adult
& $0.857 \pm 0.005$
& $0.851 \pm 0.004$
& $0.848 \pm 0.010$
& -- \\

Diabetes
& $0.585 \pm 0.004$
& $0.583 \pm 0.001$
& $0.583 \pm 0.001$
& -- \\

Census-Income
& $0.950 \pm 0.000$
& $0.949 \pm 0.002$
& $0.950 \pm 0.001$
& -- \\

CoverType
& $0.933 \pm 0.005$
& $0.935 \pm 0.001$
& $0.933 \pm 0.004$
& -- \\

\hline
IMDb
& $0.893 \pm 0.007$
& $0.892 \pm 0.008$
& $0.888 \pm 0.010$
& $0.874 \pm 0.013$ \\

TweetEval
& $0.594 \pm 0.010$
& $0.580 \pm 0.016$
& $0.577 \pm 0.020$
& $0.576 \pm 0.015$ \\

\hline
CIFAR-10
& $0.817 \pm 0.002$
& $0.807 \pm 0.009$
& $0.805 \pm 0.011$
& $0.786 \pm 0.032$ \\

DermaMNIST
& $0.751 \pm 0.006$
& $0.739 \pm 0.007$
& $0.730 \pm 0.022$
& $0.741 \pm 0.011$ \\

\hline
\end{tabular}%
}
\end{table}

Across the three optimization algorithms (i.e. GA, CHC, and PSO), similar trends are observed. RPC generally retains test accuracy close to Full HPO, although performance varies across datasets.
Matched random subsets also retain comparable accuracy, so these
results do not establish an accuracy advantage attributable to
RPC's selection objective. The contribution specific to RPC is
ranking preservation, evaluated independently on held-out
configurations in Section~\ref{sec:ranking_analysis}.

\subsection{Ranking Preservation Analysis}
\label{sec:ranking_analysis}
To evaluate the effectiveness of the proposed coreset selection strategy, we analyze the relationship between hyperparameter configuration rankings obtained from the full training dataset and those from the corresponding subsets. Ranking consensus is measured using Spearman's rank correlation coefficient ($\rho$), where higher values indicate a stronger preservation of the relative ordering of hyperparameter configurations.


\begin{table}[h]
\centering
\scriptsize

\caption{In-sample ranking fit on the $m=20$ anchor configurations used during coreset construction. Spearman rank correlation ($\rho$) compares the full-data and subset-induced hyperparameter rankings for RPC, a same-size stratified random subset, and a same-size gradient-based subset. Results are reported as mean $\pm$ standard deviation across seeds; ``--'' denotes datasets for which the gradient-based baseline is not applicable (non-differentiable Decision Tree models).}
\label{tab:ranking}

\setlength{\tabcolsep}{5pt}
\renewcommand{\arraystretch}{1.10}

\resizebox{\linewidth}{!}{%
\begin{tabular}{lcccc}
\hline
\noalign{\vskip 2pt}
\textbf{Dataset} &
\shortstack{\textbf{Subset}\\\textbf{Fraction}} &
\shortstack{\textbf{RPC}\\\textbf{Spearman}} &
\shortstack{\textbf{Random}\\\textbf{Spearman}} &
\shortstack{\textbf{Gradient-Based}\\\textbf{Spearman}} \\
\noalign{\vskip 2pt}
\hline

Dermatology
& $0.24 \pm 0.03$
& $0.863 \pm 0.040$
& $0.686 \pm 0.125$
& -- \\

Car Evaluation
& $0.18 \pm 0.06$
& $0.925 \pm 0.061$
& $0.680 \pm 0.167$
& -- \\

Abalone
& $0.24 \pm 0.04$
& $0.833 \pm 0.065$
& $0.462 \pm 0.119$
& -- \\

Adult
& $0.20 \pm 0.05$
& $0.868 \pm 0.037$
& $0.618 \pm 0.089$
& -- \\

Diabetes
& $0.21 \pm 0.04$
& $0.888 \pm 0.034$
& $0.778 \pm 0.065$
& -- \\

Census-Income
& $0.23 \pm 0.06$
& $0.905 \pm 0.036$
& $0.826 \pm 0.070$
& -- \\

CoverType
& $0.16 \pm 0.04$
& $0.974 \pm 0.012$
& $0.927 \pm 0.039$
& -- \\

\hline

IMDb
& $0.25 \pm 0.04$
& $0.786 \pm 0.053$
& $0.483 \pm 0.214$
& $0.120 \pm 0.203$ \\

TweetEval
& $0.25 \pm 0.07$
& $0.419 \pm 0.171$
& $0.287 \pm 0.206$
& $0.248 \pm 0.077$ \\

\hline

CIFAR-10
& $0.16 \pm 0.07$
& $0.970 \pm 0.011$
& $0.953 \pm 0.011$
& $0.933 \pm 0.017$ \\

DermaMNIST
& $0.23 \pm 0.06$
& $0.741 \pm 0.134$
& $0.656 \pm 0.207$
& $0.561 \pm 0.181$ \\

\hline
\end{tabular}%
}
\end{table}

Table~\ref{tab:ranking} reports ranking preservation on the anchor configurations used during the coreset construction phase. RPC achieves a higher Spearman correlation than a random subset of the same size on all 11 datasets. The advantage is particularly pronounced on datasets such as Abalone, Adult, IMDb, and TweetEval, while the differences are smaller on CIFAR-10 and CoverType, where all methods already exhibit relatively high ranking agreement. RPC also achieves higher anchor correlation than the gradient-based baseline across all four text and image datasets where the gradient-based baseline is applicable. 

RPC has higher anchor Spearman correlation than same-size Random on all 11 datasets (Table~\ref{tab:ranking}). These are in-sample measurements: the same anchor configurations contributed to RPC’s selection objective.
To assess whether these differences are systematic across datasets, a two-sided Wilcoxon signed-rank test was conducted using the dataset-level mean Spearman correlations as paired observations. For RPC versus Random, RPC obtained higher correlation on construction anchors on all 11 datasets, resulting in $W=0$, $p=9.77\times10^{-4}$, and a rank-biserial correlation of $r_{rb}=1.00$. 
These results indicate better in-sample ranking fit on the construction
anchors, but do not establish generalization or improved HPO configuration selection. Ranking generalization is evaluated separately on the held-out configurations in Table~\ref{tab:heldout_ranking}.

RPC also achieved a higher anchor Spearman correlation than GRAD-MATCH on all
four applicable datasets, yielding $W=0$, $p=0.125$, and $r_{rb}=1.00$. Although the direction and effect size consistently favor RPC, this comparison is not statistically significant. This result should be interpreted in light of the small number of applicable datasets ($n=4$).
With four paired observations, the minimum attainable $p$-value for an exact two-sided Wilcoxon signed-rank test is 0.125, even when all four differences favor the same method. Therefore, the comparison provides evidence of a consistent directional advantage, but not sufficient statistical power to establish significance.
\begin{table}[h]
\centering
\footnotesize
\caption{Spearman rank correlation ($\rho$) between the full-data and
subset-induced hyperparameter rankings on the $q=20$ held-out configurations that are disjoint from the anchor set used for coreset construction, for RPC, a same-size stratified random subset, and a same-size gradient-based subset.
Results are reported as mean $\pm$ standard deviation across seeds; ``--'' denotes datasets for which the gradient-based baseline is not applicable.
This table evaluates whether the ranking preservation observed on the anchor configurations in Table~\ref{tab:ranking} generalizes to unseen configurations (RQ3).}
\label{tab:heldout_ranking}

\setlength{\tabcolsep}{5pt}
\renewcommand{\arraystretch}{1.10}

\resizebox{\linewidth}{!}{%
\begin{tabular}{lcccc}
\hline
\noalign{\vskip 2pt}
\textbf{Dataset} &
\shortstack{\textbf{Subset}\\\textbf{Fraction}} &
\shortstack{\textbf{RPC}\\\textbf{Spearman}} &
\shortstack{\textbf{Random}\\\textbf{Spearman}} &
\shortstack{\textbf{Gradient-Based}\\\textbf{Spearman}} \\
\noalign{\vskip 2pt}
\hline

Dermatology
& $0.24 \pm 0.03$
& $0.743 \pm 0.105$
& $0.600 \pm 0.143$
& -- \\

Car Evaluation
& $0.18 \pm 0.06$
& $0.800 \pm 0.088$
& $0.676 \pm 0.154$
& -- \\

Abalone
& $0.24 \pm 0.04$
& $0.531 \pm 0.317$
& $0.493 \pm 0.181$
& -- \\

Adult
& $0.20 \pm 0.05$
& $0.685 \pm 0.137$
& $0.656 \pm 0.096$
& -- \\

Diabetes
& $0.21 \pm 0.04$
& $0.684 \pm 0.090$
& $0.574 \pm 0.150$
& -- \\

Census-Income
& $0.23 \pm 0.06$
& $0.668 \pm 0.172$
& $0.621 \pm 0.183$
& -- \\

CoverType
& $0.16 \pm 0.04$
& $0.952 \pm 0.016$
& $0.955 \pm 0.023$
& -- \\

\hline

IMDb
& $0.25 \pm 0.04$
& $0.514 \pm 0.104$
& $0.294 \pm 0.338$
& $-0.200 \pm 0.174$ \\

TweetEval
& $0.25 \pm 0.07$
& $0.073 \pm 0.180$
& $-0.004 \pm 0.169$
& $-0.067 \pm 0.219$ \\

\hline

CIFAR-10
& $0.16 \pm 0.07$
& $0.873 \pm 0.102$
& $0.896 \pm 0.016$
& $0.774 \pm 0.073$ \\

DermaMNIST
& $0.23 \pm 0.06$
& $0.490 \pm 0.053$
& $0.516 \pm 0.231$
& $0.465 \pm 0.244$ \\

\hline
\end{tabular}%
}
\end{table}

Table~\ref{tab:heldout_ranking} provides a more stringent evaluation by measuring ranking preservation on configurations that were not used during coreset construction. As expected, the correlations are generally lower than those obtained on the anchor configurations, since the coreset was explicitly optimized with respect to the anchor ranking rather than the unseen configurations. Nevertheless, RPC retains higher held-out correlation than Random on 8 of the 11 datasets, including six of the seven tabular datasets and both text datasets.

The dataset-level Wilcoxon signed-rank test confirms that the advantage over Random extends beyond the anchor configurations. RPC versus Random yields $W=6$, $p=0.014$, and $r_{rb}=0.82$, indicating a statistically significant and substantial overall advantage in held-out ranking preservation. These results provide evidence that the ranking information captured during coreset construction generalizes to configurations that were not directly involved in the selection objective, although the strength of this generalization varies across datasets.

For the gradient-based comparison, RPC achieves higher held-out Spearman correlation on all four applicable datasets, yielding $W=0$, $p=0.125$,
and $r_{rb}=1.00$. Although the direction and effect size consistently favor
RPC, the comparison is not statistically significant because the gradient-based baseline applies only to four (n = 4) differentiable datasets. With four paired observations, the minimum attainable $p$-value for an exact two-sided Wilcoxon signed-rank test is 0.125. Notably, RPC maintains a positive mean held-out Spearman correlation across all 11 datasets, whereas the gradient-based baseline produces negative mean correlations on IMDb and TweetEval. This further indicates that the ranking-preserving subsets retain a positive association with the full-data configuration ranking across all evaluated datasets.

The held-out results on the image datasets show relatively close performance between RPC and Random, with Random achieving a slightly higher mean Spearman correlation in both cases. However, the observed differences are small relative
to the run-to-run variability and should therefore be interpreted cautiously.
These results suggest that held-out ranking preservation may vary across datasets and experimental seeds rather than indicating a systematic disadvantage of RPC on image datasets. Importantly, RPC maintains a positive held-out Spearman correlation across all 11 datasets, indicating that the subset-induced rankings remain positively associated with the corresponding full-data rankings even on unseen configurations. Overall, the aggregate held-out analysis provides evidence that the ranking-preservation advantage of RPC extends beyond the anchor configurations, although the magnitude of this advantage varies across datasets. Moreover, the overall held-out comparison remains statistically significant across all 11 datasets ($W=6$, $p=0.014$), even after including the two image datasets. We therefore consider the image results an informative boundary for the current framework. Further work could examine alternative anchor-selection strategies or incorporate ranking stability into the selection objective. 

Overall, the results distinguish between two aspects of ranking preservation. RPC achieves stronger in-sample ranking fit than same-size
random sampling on the construction anchors. More importantly, the held-out analysis shows that this advantage extends to unseen configurations at the aggregate dataset level, demonstrating that the selected subsets capture ranking information beyond the anchor set. At the same time, the weaker held-out behavior on the image datasets indicates that this generalization is not uniform across all modalities and should be interpreted conservatively.

To examine whether the subset-induced rankings identify promising
configurations, we additionally evaluate top-$k$ regret on the same
$q=20$ held-out configurations, for $k\in\{1,3,5\}$. For each dataset
and seed, each method ranks these configurations using its subset-based
validation scores. 

Let $\mathsf{P}$ denote the set of 20 held-out configurations,
and let $\mathsf{Q}_k(M)\subseteq\mathsf{P}$ denote the top-$k$
configurations selected using method $M$'s subset-based
validation scores, where $M$ is RPC or Random.
Let $a_D(c)$ denote the recorded validation accuracy of
configuration $c$ trained on the full training data.
The top-$k$ regret is
\begin{equation}
\ell_k(M,\mathsf{P})
=
\max_{c \in \mathsf{P}} a_D(c)
-
\max_{c \in \mathsf{Q}_k(M)} a_D(c).
\label{eq:heldout_regret}
\end{equation}
Lower regret indicates that the shortlist contains a configuration
closer in full-data validation accuracy to the best of the 20 held-out
candidates. When subset scores tie at the shortlist boundary, we average
regret over all equally likely selections among the tied configurations.
Regret is first averaged across seeds within each dataset; aggregate
results then give each dataset equal weight. This retrospective analysis
uses the full-data scores already recorded for the held-out candidates
and evaluates shortlist quality rather than the final test accuracy
achieved by GA, CHC, or PSO.
RPC has lower aggregate mean regret than Random at all three
shortlist sizes. The difference is statistically significant for
$k=5$ ($p=0.0078$), but not for $k=1$ or $k=3$. These results
suggest an advantage in identifying a five-configuration shortlist,
rather than establishing improved final HPO test accuracy.
\begin{table}[htbp]
\centering
\caption{Top-$k$ regret on 20 held-out configurations.
Values are averaged across seeds within each dataset and then
equally across 11 datasets. Lower is better. The $p$-values
are from two-sided Wilcoxon signed-rank tests on paired
dataset-level mean regrets ($n=11$), comparing RPC with Random,
and are unadjusted for the three comparisons.}
\label{tab:heldout_regret}
\begin{tabular}{lccc}
\hline
Shortlist & RPC & Random & $p$-value \\
\hline
Top-1 & 0.00818 & 0.01328 & 0.0547 \\
Top-3 & 0.00350 & 0.00518 & 0.5703 \\
Top-5 & 0.00139 & 0.00294 & 0.0078 \\
\hline
\end{tabular}
\end{table}

\subsection{Computational Efficiency and Break-Even}
\label{sec:breakeven}

\begin{figure}[H]
\centering
\includegraphics[width=\textwidth]{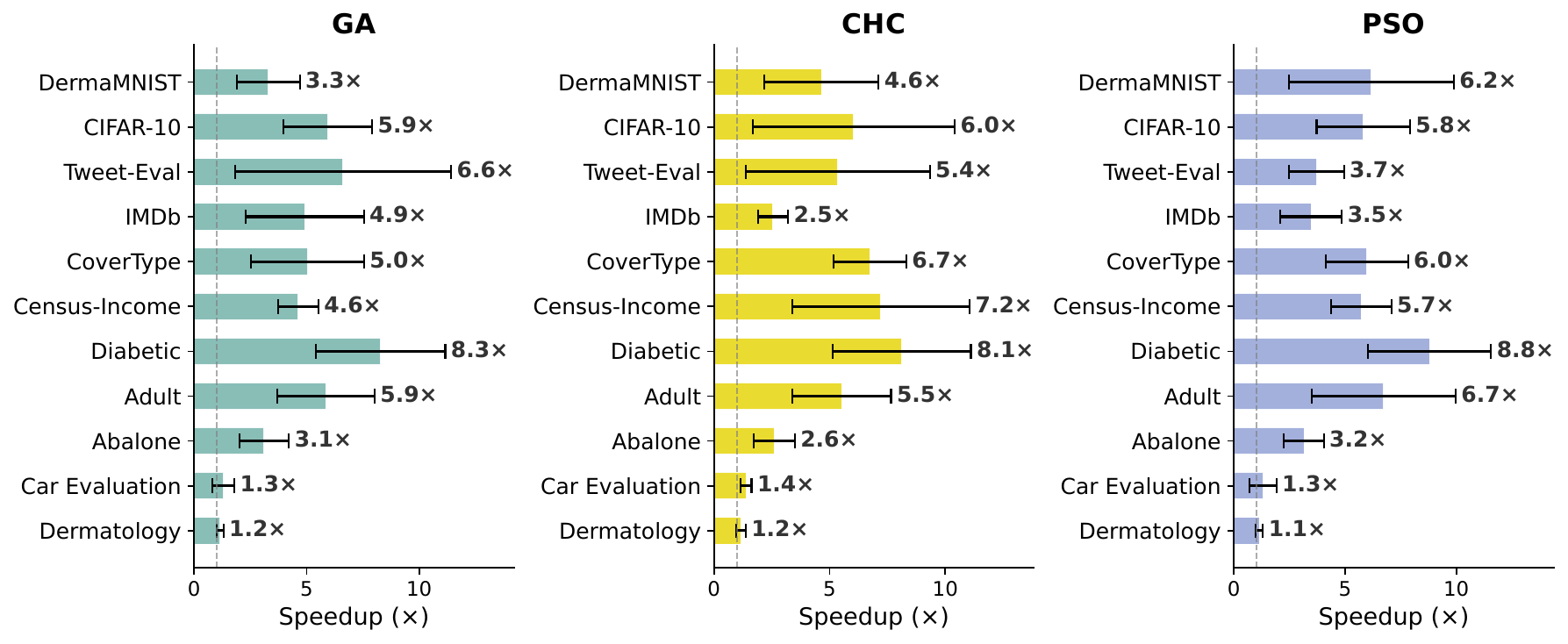}
\caption{
Comparison of hyperparameter optimization runtime speedup achieved by the proposed ranking-preserving coreset across different datasets and optimization algorithms. Speedup is computed as the ratio of Full HPO time to Coreset HPO time. Error bars represent standard deviation across experimental runs.
}
\label{fig:speedup}
\end{figure}

The primary goal of the proposed framework is to reduce computational cost during hyperparameter optimization. Table~\ref{tab:runtime} reports the runtime comparison between full-dataset HPO and coreset-based HPO. The corresponding computational gains are illustrated in Figure~\ref{fig:speedup}, which compares the speedups achieved across datasets and optimization algorithms. The corresponding speedup is computed as the ratio of full-dataset HPO time to coreset HPO time. 

The results indicate that once the coreset has been constructed, HPO on the selected subset substantially reduces configuration evaluation time across tabular, text, and image datasets. This reduction is especially visible on computationally expensive datasets, where repeated model training dominates the HPO cost.

\begin{table}[H]
\centering
\footnotesize
\caption{HPO runtime and speedup comparison. Speedups are calculated within each seed and then summarized as mean $\pm$ standard deviation}
\label{tab:runtime}
\begin{tabular}{llccc}
\hline
\textbf{Dataset} & \textbf{Optimizer} & $T_{\text{full}}$ \textbf{(s)} & $T_{\text{core}}$ \textbf{(s)} & 
Speedup $\mathbf{(\times)}$ \\
\hline
Dermatology & GA & $0.27 \pm 0.02$ & $0.23 \pm 0.02$ & $1.16 \pm 0.15$ \\
 & CHC & $0.74 \pm 0.10$ & $0.65 \pm 0.10$ & $1.15 \pm 0.21$ \\
 & PSO & $0.27 \pm 0.04$ & $0.24 \pm 0.03$ & $1.14 \pm 0.17$ \\
 \hline
Car Evaluation & GA & $0.37 \pm 0.12$ & $0.30 \pm 0.06$ & $1.31 \pm 0.49$ \\
 & CHC & $0.93 \pm 0.10$ & $0.68 \pm 0.10$ & $1.39 \pm 0.23$ \\
 & PSO & $0.40 \pm 0.11$ & $0.34 \pm 0.10$ & $1.32 \pm 0.60$ \\
\hline
Abalone & GA & $1.92 \pm 0.43$ & $0.66 \pm 0.17$ & $3.11 \pm 1.10$ \\
 & CHC & $3.41 \pm 1.01$ & $1.35 \pm 0.27$ & $2.60 \pm 0.89$ \\
 & PSO & $1.58 \pm 0.67$ & $0.50 \pm 0.17$ & $3.15 \pm 0.90$ \\
\hline
Adult & GA & $10.71 \pm 2.57$ & $1.91 \pm 0.39$ & $5.86 \pm 2.17$ \\
 & CHC & $28.48 \pm 4.05$ & $5.75 \pm 1.89$ & $5.52 \pm 2.13$ \\
 & PSO & $13.76 \pm 3.56$ & $2.30 \pm 0.71$ & $6.73 \pm 3.25$ \\
\hline
Diabetes & GA & $710.68 \pm 109.38$ & $90.99 \pm 23.03$ & $8.29 \pm 2.87$ \\
 & CHC & $1608.11 \pm 707.41$ & $199.28 \pm 69.52$ & $8.12 \pm 2.99$ \\
 & PSO & $937.92 \pm 89.89$ & $114.48 \pm 32.06$ & $8.80 \pm 2.77$ \\
 \hline
Census-Income & GA & $43.07 \pm 3.85$ & $9.46 \pm 1.21$ & $4.63 \pm 0.89$ \\
 & CHC & $128.67 \pm 40.94$ & $20.27 \pm 8.34$ & $7.22 \pm 3.83$ \\
 & PSO & $58.75 \pm 5.41$ & $10.59 \pm 1.81$ & $5.72 \pm 1.37$ \\
\hline
CoverType & GA & $294.63 \pm 60.52$ & $73.37 \pm 41.76$ & $5.04 \pm 2.51$ \\
 & CHC & $1121.62 \pm 231.29$ & $172.79 \pm 48.65$ & $6.74 \pm 1.57$ \\
 & PSO & $507.25 \pm 116.72$ & $88.97 \pm 28.66$ & $5.99 \pm 1.84$ \\
\hline
IMDb & GA & $829.13 \pm 326.16$ & $182.51 \pm 37.10$ & $4.91 \pm 2.63$ \\
 & CHC & $1356.48 \pm 242.09$ & $550.01 \pm 133.27$ & $2.55 \pm 0.65$ \\
 & PSO & $907.00 \pm 434.68$ & $273.43 \pm 91.11$ & $3.47 \pm 1.39$ \\
\hline
TweetEval  & GA & $330.29 \pm 115.98$ & $69.75 \pm 40.65$ & $6.62 \pm 4.81$ \\
 & CHC & $1113.40 \pm 1020.47$ & $248.78 \pm 120.44$ & $5.36 \pm 3.97$ \\
 & PSO & $349.81 \pm 114.66$ & $106.78 \pm 61.40$ & $3.72 \pm 1.23$ \\
\hline
CIFAR-10 & GA & $2559.70 \pm 355.94$ & $467.03 \pm 190.19$ & $5.95 \pm 1.97$ \\
 & CHC & $4091.20 \pm 2071.77$ & $1262.19 \pm 1532.50$ & $6.05 \pm 4.36$ \\
 & PSO & $2767.52 \pm 386.22$ & $502.61 \pm 129.53$ & $5.82 \pm 2.10$ \\
\hline
DermaMNIST & GA & $1012.29 \pm 27.33$ & $342.06 \pm 123.08$ & $3.29 \pm 1.40$ \\
 & CHC & $1807.70 \pm 412.64$ & $438.24 \pm 169.23$ & $4.65 \pm 2.47$ \\
 & PSO & $1327.84 \pm 552.73$ & $248.14 \pm 147.93$ & $6.18 \pm 3.71$ \\
\hline
\end{tabular}
\end{table}

However, the per-run speedup defined in Equation~\ref{eq:speedup}
does not account for the initial coreset construction overhead.
We therefore also report the construction cost and the break-even
point defined in Equation~\ref{eq:breakeven}, which quantifies
the number of repeated HPO executions required to recover
that cost.

Let $T_{\text{sample}}$ denote the one-time coreset construction time, $T_{\text{full}}$ denote the HPO runtime on the full training dataset, and $T_{\text{core}}$ denote the corresponding HPO runtime on the selected coreset. A smaller value of $N_{\text{BE}}$ indicates that the proposed framework becomes computationally beneficial after fewer optimization runs.

The proposed coreset selection stage is designed to be independent of the downstream HPO optimizer, as the subset is selected solely based on the ranking-preservation objective with respect to anchor configurations. It does not use optimizer-specific information from GA, CHC, or PSO. To evaluate this optimizer-independent setting, we select one coreset for each dataset--seed split and reuse the same selected subset across GA, CHC, and PSO. The coreset selection algorithm is presented in Algorithm~\ref{alg:active_sampling}.

To examine whether the effect of using the selected coreset varies across the downstream optimizers, we compared the predictive-performance differences between full-data HPO and coreset-based HPO reported in
Tables~\ref{tab:ga_accuracy}--\ref{tab:pso_accuracy} for GA, CHC, and PSO. A Friedman test conducted over the 11 datasets revealed no significant difference in the Full--RPC accuracy gap across the three optimizers ($\chi^{2}=2.39$, $p=0.303$). 

We further examined whether the dataset-level accuracy differences followed similar patterns across optimizers. The gaps were strongly correlated between GA and CHC (Spearman $\rho=0.89$, $p<0.001$) and moderately correlated between CHC and PSO ($\rho=0.67$, $p=0.024$). A similar positive relationship was observed between GA and PSO ($\rho=0.60$, $p=0.054$), although this comparison did not reach the $0.05$ significance level. Thus, datasets showing a relatively larger performance gap under one optimizer generally showed a larger gap under the others as well.

The observed accuracy-gap patterns were broadly similar across the three evaluated optimizers, and the Friedman test did not detect a significant difference between them.


Since the same selected coreset is reused across GA, CHC, and PSO under each dataset--seed split, the primary break-even analysis is performed at the complete optimization-cycle level, where one cycle consists of one run of each of the three optimizers.
The cycle-level break-even point is therefore defined as:

\[
N_{\text{BE\_cycle}}
=
\frac{
T_{\text{sample}}
}{
(T_{\text{fullGA}}+T_{\text{fullCHC}}+T_{\text{fullPSO}})
-
(T_{\text{CoreGA}}+T_{\text{coreCHC}}+T_{\text{corePSO}})
}.
\]

Table~\ref{tab:end_to_end_breakeven} reports the end-to-end runtime and cycle-level
break-even analysis relative to repeated Full HPO.
At a fixed subset budget, RPC offers no inherent computational
advantage over same-size random sampling for configuration
evaluation: both methods train on the same number of instances,
while RPC incurs additional coreset construction overhead.
The benefit sought through this additional selection cost is
improved ranking fidelity, evaluated on held-out configurations
in Table~\ref{tab:heldout_ranking}. The break-even analysis therefore quantifies how
many repeated Full HPO runs are needed to recover RPC's
construction cost; it does not establish a computational
advantage over Random.

The break-even results show that the computational benefit of the proposed
framework depends on the relationship between the one-time coreset construction cost and the savings obtained during subsequent HPO runs. For very small tabular datasets such as Dermatology (366 instances) and Car Evaluation
(1,728 instances), full-data HPO is already inexpensive relative to the coreset construction stage: a complete full-data GA--CHC--PSO cycle requires only 1.29~s and 1.71~s, respectively, so the construction cost is amortized after approximately 35.69 and 15.93 complete optimization cycles.
This pattern reflects the intended operating regime of coreset-based HPO acceleration. As established in Section~\ref{sec:intro}, the computational burden that motivates HPO acceleration comes specifically from large-scale, high-dimensional, and deep-learning workloads, where repeated model training is expensive. A problem whose full HPO cycle already finishes in about a second, as in Dermatology and Car Evaluation, sits outside that regime entirely.

For a single-optimizer HPO run, RPC provides net runtime savings only when 
\(T_{\mathrm{sample}}+T_{\mathrm{core}}<T_{\mathrm{full}}\). Based on the reported mean runtimes in Table~\ref{tab:optimizer_level_breakeven}, this condition holds only for CHC on Diabetes, Census-Income, and CoverType. All other evaluated combinations, including every text and image setting, require reuse to recover construction cost. By contrast, one optimization cycle in Table~\ref{tab:end_to_end_breakeven} comprises three runs—one each of GA, CHC, and PSO.

At the cycle level, break-even is reached within a single optimization cycle for Adult, Diabetes, Census-Income, and CoverType ($N_{\mathrm{BE\text{-}cycle}}=0.62$, $0.31$, $0.51$, and $0.36$, respectively), and remains low between 0.78 and 2.52 cycles for every text and image dataset evaluated, including the computationally demanding
CIFAR-10 and DermaMNIST benchmarks that require training a ResNet18 model. These results indicate that RPC recovers its construction cost after fewer repeated HPO cycles on the evaluated larger tabular, text, and image datasets than on Dermatology and
Car Evaluation. Its practical advantage therefore lies in settings where savings from repeated configuration evaluations offset the initial construction cost. For very small datasets, full-data HPO is already inexpensive, so many more repeated
cycles are needed before RPC provides an overall runtime benefit.

The behavior becomes substantially more favorable as the cost of full-data HPO increases. Abalone reaches break-even after approximately $2.16$ optimization cycles. For example, on Diabetes, the full optimization cycle requires $3256.70$~s, compared with $404.75$~s using the coreset; even after including the $890.76$~s construction cost, the first coreset-based cycle remains substantially less expensive than the corresponding full-data cycle.

\begin{table}[H] 
\centering 
\footnotesize 

\caption{Cycle-level end-to-end break-even analysis. $T_{\mathrm{sample}}$ is the one-time coreset construction time (s); $T_{\mathrm{full\text{-}cycle}}$
and $T_{\mathrm{core\text{-}cycle}}$ are the summed full-data and
coreset-based HPO runtimes (s) for one complete GA--CHC--PSO cycle;
$T_{\mathrm{sample}} + T_{\mathrm{core\text{-}cycle}}$ is the total coreset-based cost of the first cycle, including construction, and $N_{\mathrm{BE\text{-}cycle}}$ is the number of complete cycles required for the cumulative coreset-based cost to fall below the cumulative full-data cost, as defined above. Values of $N_{\mathrm{BE\text{-}cycle}}<1$ indicate that the construction cost is recovered within the first optimization cycle.} 

\label{tab:end_to_end_breakeven} 
\setlength{\tabcolsep}{5pt}
\renewcommand{\arraystretch}{1.10}

\resizebox{\linewidth}{!}{%
\begin{tabular}{lccccc} \hline \textbf{Dataset} & $T_{\text{sample}}$ & $T_{\text{full-cycle}}$ & $T_{\text{core-cycle}}$ & $T_{\text{sample}}+T_{\text{core-cycle}}$ & $N_{\text{BE-cycle}}$ \\ \hline Dermatology & $5.67$ & $1.29$ & $1.13$ & $6.80$ & $35.69$ \\ Car Evaluation & $6.25$ & $1.71$ & $1.31$ & $7.56$ & $15.93$ \\ Abalone & $9.52$ & $6.91$ & $2.51$ & $12.03$ & $2.16$ \\ Adult & $26.75$ & $52.95$ & $9.96$ & $36.70$ & $0.62$ \\ Diabetes & $890.76$ & $3256.70$ & $404.75$ & $1295.51$ & $0.31$ \\ Census-Income & $97.56$ & $230.48$ & $40.32$ & $137.87$ & $0.51$ \\ CoverType & $567.70$ & $1923.50$ & $335.13$ & $902.83$ & $0.36$ \\ IMDb & $3427.26$ & $3092.62$ & $1005.95$ & $4433.20$ & $1.64$ \\ TweetEval  & $1064.54$ & $1793.50$ & $425.32$ & $1489.86$ & $0.78$ \\ CIFAR-10 & $13229.94$ & $9418.41$ & $2231.83$ & $15461.76$ & $1.84$ \\ DermaMNIST & $7854.88$ & $4147.84$ & $1028.44$ & $8883.32$ & $2.52$ \\ \hline \end{tabular} 
}
\end{table} 
The optimizer-level break-even values are provided in~\ref{app:optimizer_breakeven}, Table~\ref{tab:optimizer_level_breakeven}, to avoid overcrowding the main results section.

\begin{figure}[h]
\centering
\includegraphics[width=\textwidth]{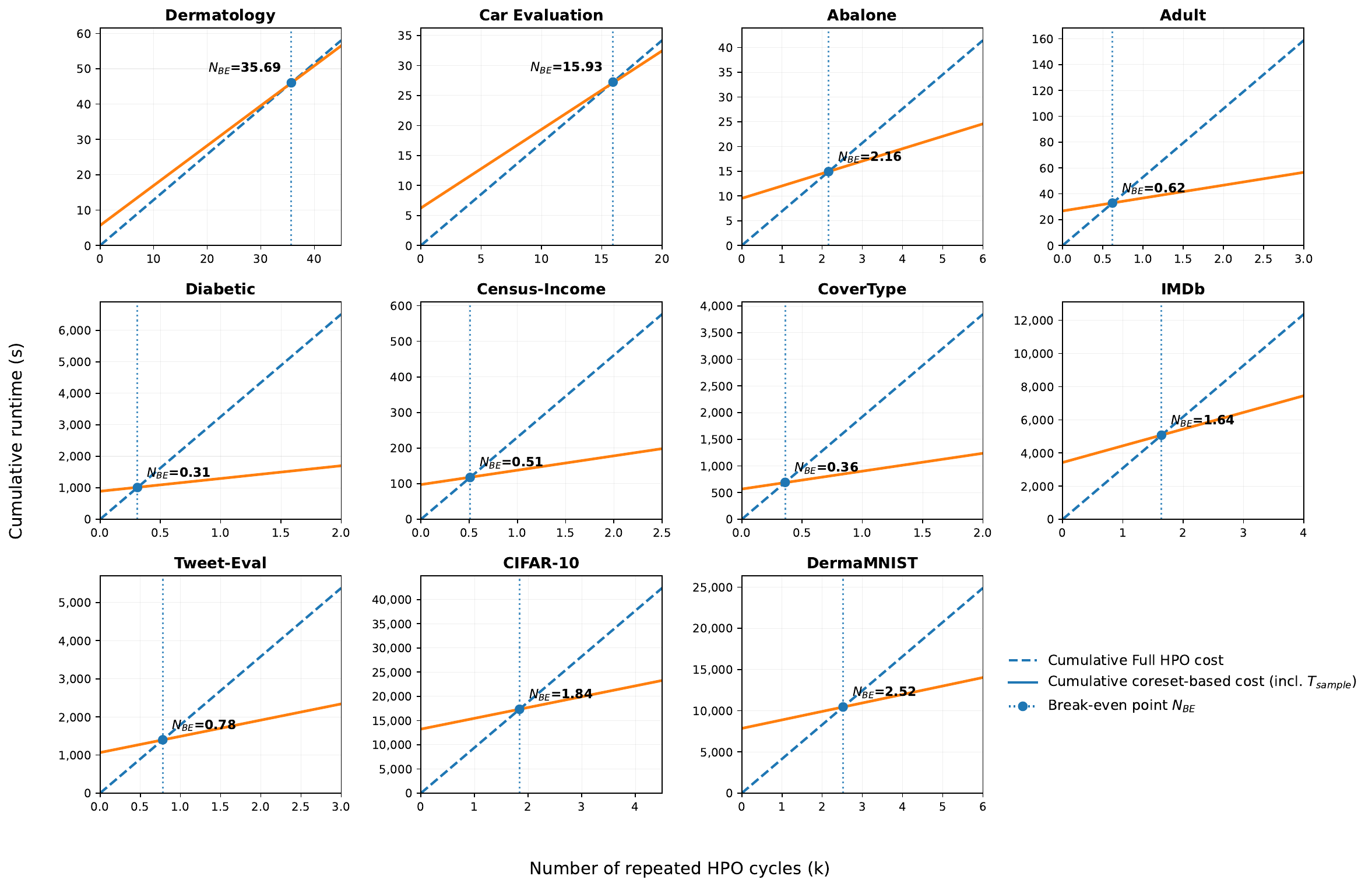}
\caption{Cumulative runtime break-even analysis for representative datasets. The crossover point indicates the number of repeated HPO cycles after which the cumulative cost of coreset construction and coreset-based HPO becomes lower than that of repeated full-data HPO.}
\label{fig:breakeven}
\end{figure}

For the text datasets, TweetEval Sentiment reaches break-even within the first complete optimization cycle ($N_{\text{BE-cycle}}=0.78$), whereas IMDb requires approximately $1.64$ cycles. The image datasets also exhibit practical amortization despite their comparatively high coreset construction costs: CIFAR-10 reaches break-even after approximately $1.84$ cycles and DermaMNIST after $2.52$ cycles. Thus, for computationally expensive neural-network HPO, the initial construction overhead can be recovered after only a few repeated optimization cycles. Figure~\ref{fig:breakeven} illustrates the cumulative runtime behavior for representative datasets.

Overall, the end-to-end analysis provides a more conservative assessment of computational efficiency than per-run HPO speedup alone. The proposed method is less attractive for very small datasets where full-data HPO is already inexpensive, but it becomes increasingly advantageous when configuration evaluation is costly or when the selected coreset is reused across multiple optimization runs. These results, therefore, support its use in repeated model selection, AutoML, and other computationally intensive HPO settings.

\subsection{Discussion}

RPC is a coreset-selection objective that preserves hyperparameter ranking while penalizing subset size. It does not aim to improve predictive accuracy over Full-data HPO or same-size random sampling. This work further evaluates the held-out ranking agreement to generalize beyond the construction anchors, while selected-configuration accuracy and top-$k$ regret assess the practical consequences for model selection. The experimental results provide evidence that the proposed ranking-preserving coreset framework can reduce the training data used during hyperparameter optimization while retaining much of the predictive performance using full-data HPO. Across the 33 dataset--optimizer combinations evaluated using GA, CHC, and PSO, the mean absolute difference between Full HPO and RPC HPO is approximately $0.73$ percentage points. In 30 of the 33 comparisons, the difference remains within two percentage points, while the largest observed absolute difference is approximately $2.4$ percentage points. These results indicate that RPC can identify competitive configurations using reduced training data, although its selected-configuration accuracy is not equivalent to that of full-data HPO in every case.
Second, the ranking-preservation results distinguish in-sample fit from generalization. RPC achieves a higher Spearman correlation than same-size random subsets across all 11 datasets for the anchor configurations (Table~\ref{tab:ranking}), which contribute to the coreset construction objective. The dataset-level Wilcoxon signed-rank test ($W=0$, $p=9.77\times10^{-4}$, $r_{rb}=1.00$) therefore describes the in-sample fit to the construction anchors and does not establish generalization or improved configuration selection. Generalization is evaluated separately using disjoint held-out configurations.  

The held-out configuration analysis provides a more stringent assessment of this property. Although correlations are lower than for the anchor configurations, RPC achieves a higher held-out Spearman correlation than Random on 8 of the 11 datasets, and the overall difference remains statistically significant ($W=6$, $p=0.0137$, $r_{rb}=0.82$). This indicates that the ranking information captured during coreset construction extends beyond the configurations explicitly used by the selection objective. However, this generalization is not uniform across all datasets. In particular, RPC is slightly below Random on the two image datasets in the held-out analysis. RPC’s mean held-out correlation was 0.873 on CIFAR-10 and 0.490 on DermaMNIST, indicating different degrees of ranking preservation within the image setting. One possible explanation is anchor dependence: RPC selects examples to preserve the construction-anchor rankings, which may not adequately represent how unseen configurations respond to reduced training data. Random sampling does not adapt to these anchor rankings and may therefore perform better when this dependence limits generalization. Differences in training dynamics across ResNet18 configurations could also contribute, but these explanations remain unverified. Given the evaluation’s restriction to two image datasets, three seeds, and one architecture, broader image-task generalization cannot be established. Strong anchor fit should therefore be supplemented by held-out ranking evaluation before relying on RPC in a new image-classification setting.

Third, the top-$k$ regret analysis provides a complementary assessment of configuration selection. RPC has lower aggregate mean regret than Random at $k=1$, $3$, and $5$, but the paired dataset-level comparisons are statistically significant only at $k=5$
($p=0.0078$). These findings suggest improved five-configuration shortlist quality, but do not establish a consistent advantage in selecting a single configuration or improving final HPO test accuracy. The analysis retrospectively uses recorded full-data validation scores; successive-halving and transfer to another model family remain potential application regimes requiring direct evaluation.

Furthermore, the computational efficiency results show that reducing the training data used for configuration evaluation can substantially decrease HPO-stage runtime. The observed HPO-stage speedups range from approximately $1.14\times$ to $8.80\times$, although these values do not by themselves account for the one-time cost of constructing the coreset. The end-to-end break-even analysis provides a more conservative assessment. For very small datasets such as Dermatology and Car Evaluation, full-data HPO is already inexpensive, and many repeated optimization cycles are required before the coreset construction cost is recovered. In contrast, Adult, Diabetes, Census-Income, CoverType, and TweetEval Sentiment reach break-even within the first complete GA--CHC--PSO cycle, while IMDb, CIFAR-10, and DermaMNIST require only a small number of repeated cycles. These results indicate that the proposed framework is most attractive when configuration evaluation itself is computationally expensive or when the selected coreset can be reused across multiple HPO executions.

Importantly, the predictive performance and efficiency results should be considered together. RPC does not eliminate the performance gap relative to Full HPO, nor does it provide a runtime advantage on every dataset after accounting for selection cost. Rather, it offers a trade-off: an average reduction of less than one percentage point across the evaluated dataset--optimizer combinations, in exchange for substantially cheaper repeated configuration evaluations. The usefulness of this trade-off therefore depends on the underlying dataset, model-training cost, and the number of HPO runs over which the one-time selection cost can be amortized.

The gradient-based baseline does not consistently match RPC's ranking preservation. Further work could examine whether more diverse anchor sets or a ranking-stability objective improves generalization to unseen configurations.

\subsection{Limitations}
\label{sec:limitations}
Although the framework shows benefits in several evaluated settings, it has limitations. First, the coreset selection stage introduces an initial computational overhead,
which is most beneficial when the selected coreset is reused across multiple HPO
runs or model-development cycles. Second, the current framework relies on a fixed
set of anchor hyperparameter configurations to estimate the full-data ranking; therefore, the quality of the selected coreset may depend on the diversity of
these anchor configurations. Third, image experiments are computationally more
expensive than tabular experiments because each configuration evaluation requires training a CNN. 
Fourth, RPC’s held-out ranking advantage did not extend to CIFAR-10 and DermaMNIST, where Random achieved slightly higher mean correlations. This indicates that strong anchor fit does not guarantee ranking preservation on unseen configurations; the underlying causes remain unverified. The image evaluation is also limited to two datasets, three seeds per dataset, and one architecture (ResNet18), restricting conclusions about run-to-run variability and generalization to other image tasks and model families. Broader evaluation with additional datasets, seeds, and architectures is needed. Finally, the sensitivity analysis evaluates anchor count under the chosen selection procedure but does not separately assess sensitivity to alternative anchor sets of the same size. The present study evaluates classification tasks only; future work should extend the framework to regression, multi-label learning, and larger transformer-based models.

\section{Conclusion}
\label{sec:conclusion}

This paper proposed a ranking-preserving coreset selection framework for efficient hyperparameter optimization. The central idea is to construct a compact coreset that preserves the relative ordering of hyperparameter configurations observed on the full training dataset. By directly optimizing ranking preservation, the selected subset provides a reduced-data approximation of the configuration ordering required for HPO, thereby lowering the cost of repeated model evaluations.

The proposed framework was evaluated on 11 benchmark datasets, including 7 tabular, 2 text, and 2 image datasets. Experiments were conducted using Decision Trees for tabular data, TF-IDF with a PyTorch MLP for text data, and ResNet18 for image data. Three evolutionary optimization algorithms---GA, CHC, and PSO---were used to examine the behavior of the proposed approach across different HPO strategies.

On 20 held-out configurations, RPC achieved higher mean Spearman correlation than same-size random on 8 of 11 datasets (dataset-level Wilcoxon signed-rank test, $p=0.014$). This provides evidence that ranking preservation extends beyond the construction anchors, although the advantage is not uniform: Random obtained slightly higher mean correlations on both image datasets. Correlations measured on the construction anchors describe in-sample fit and are not evidence of ranking
generalization or improved final configuration selection.

In terms of predictive performance, RPC HPO remains close to full-data HPO across most dataset--optimizer combinations, with an average absolute accuracy difference of less than one percentage point. At the same time, reducing the training data substantially decreases the cost of individual HPO evaluations, producing HPO-stage speedups ranging from $1.14\times$ to $8.80\times$. The end-to-end break-even analysis further shows that the practical benefit depends on the cost of coreset construction and the number of repeated optimization runs. The framework is therefore most advantageous for computationally expensive or repeated HPO settings, whereas the construction overhead may outweigh the savings for very small datasets.

Overall, the findings support ranking preservation as a useful criterion for constructing subsets for hyperparameter optimization. The proposed framework provides a practical trade-off among ranking fidelity, predictive performance, and computational cost, while also identifying limitations in ranking preservation on held-out
hyperparameter configurations for image datasets.
Future work will investigate adaptive weighting in the coreset fitness function, broader and more stable anchor coverage, larger-scale transformer-based models, regression and multi-label tasks, and integration with automated machine learning pipelines.
\\

\section*{Declaration of Generative AI and AI-Assisted Technologies in the Manuscript Preparation Process}

During the preparation of this work, the authors used ChatGPT by OpenAI to support language editing, wording refinement, manuscript clarity, and code generation and debugging. Grammarly was used for grammar, spelling, and sentence-level language refinement. The authors reviewed, verified, and edited all generated outputs as needed and took full responsibility for the accuracy, integrity, and content of the published article.


\clearpage
\appendix
\section{Dataset Class Distributions}
\label{app:class_distributions}
\begingroup
\footnotesize
\setlength{\tabcolsep}{5pt}
\renewcommand{\arraystretch}{1.3}
\begin{longtable}{
    @{}
    p{0.13\textwidth}
    r
    p{0.68\textwidth}
    @{}
}
\caption{Dataset class distributions for the loaded tabular datasets, selected 10,000-example text datasets, and complete image datasets, including their official test partitions. Tabular distributions are reported before rare-class merging during stratified splitting. Class IDs correspond to the labels encoded by the data loaders. Entries are shown as class ID: count (percentage), with percentages rounded to two decimal places.}\\
\label{tab:class_distributions}\\
\hline
\textbf{Dataset} & \textbf{Instances} &
\textbf{Class counts and percentages} \\
\hline
\endfirsthead

\multicolumn{3}{l}{\textit{Table~\ref{tab:class_distributions}
continued}}\\
\hline
\textbf{Dataset} & \textbf{Instances} &
\textbf{Class counts and percentages} \\
\hline
\endhead

\hline
\multicolumn{3}{r}{\textit{Continued on next page}}\\
\endfoot

\hline
\endlastfoot

Dermatology & 366 &
0: 112 (30.60\%);
1: 61 (16.67\%);
2: 72 (19.67\%);
3: 49 (13.39\%);
4: 52 (14.21\%);
5: 20 (5.46\%). \\

Car Evaluation & 1,728 &
0: 384 (22.22\%);
1: 69 (3.99\%);
2: 1,210 (70.02\%);
3: 65 (3.76\%). \\

Abalone & 4,177 &
0: 1 (0.02\%);
1: 634 (15.18\%);
2: 487 (11.66\%);
3: 267 (6.39\%);
4: 203 (4.86\%);
5: 126 (3.02\%);
6: 103 (2.47\%);
7: 67 (1.60\%);
8: 58 (1.39\%);
9: 42 (1.01\%);
10: 32 (0.77\%);
11: 1 (0.02\%);
12: 26 (0.62\%);
13: 14 (0.34\%);
14: 6 (0.14\%);
15: 9 (0.22\%);
16: 2 (0.05\%);
17: 1 (0.02\%);
18: 1 (0.02\%);
19: 2 (0.05\%);
20: 1 (0.02\%);
21: 15 (0.36\%);
22: 57 (1.36\%);
23: 115 (2.75\%);
24: 259 (6.20\%);
25: 391 (9.36\%);
26: 568 (13.60\%);
27: 689 (16.50\%). \\

Adult & 32,561 &
0: 24,720 (75.92\%);
1: 7,841 (24.08\%). \\

Diabetic & 101,766 &
0: 11,357 (11.16\%);
1: 35,545 (34.93\%);
2: 54,864 (53.91\%). \\

Census-Income & 199,523 &
0: 187,141 (93.79\%);
1: 12,382 (6.21\%). \\

CoverType & 581,012 &
0: 211,840 (36.46\%);
1: 283,301 (48.76\%);
2: 35,754 (6.15\%);
3: 2,747 (0.47\%);
4: 9,493 (1.63\%);
5: 17,367 (2.99\%);
6: 20,510 (3.53\%). \\

IMDb & 10,000 &
0: 5,000 (50.00\%);
1: 5,000 (50.00\%). \\

TweetEval & 10,000 &
0: 1,899 (18.99\%);
1: 4,588 (45.88\%);
2: 3,513 (35.13\%). \\

CIFAR-10 & 60,000 &
Classes 0--9: 6,000 instances (10.00\%) per class. \\

DermaMNIST & 10,015 &
0: 327 (3.27\%);
1: 514 (5.13\%);
2: 1,099 (10.97\%);
3: 115 (1.15\%);
4: 1,113 (11.11\%);
5: 6,705 (66.95\%);
6: 142 (1.42\%). \\

\end{longtable}
\endgroup

\clearpage
\section{Supplementary Diagnostics}
\label{app:sensitivity}
\subsection{Sensitivity to the Number of Anchor Configurations}
\label{app:anchor_sensitivity}
\begin{table}[htbp]
\centering
\small
\caption{Dataset-level sensitivity of RPC held-out ranking preservation to construction-anchor count.
Spearman correlations compare full-data and RPC rankings on the same fixed 20 held-out configurations within each dataset--seed combination. Values are mean $\pm$ standard
deviation across seeds.}
\label{tab:sensitivity_dataset_detail}
\begin{tabular}{lcccc}
\hline
Dataset & Seeds & 10 anchors & 20 anchors & 30 anchors \\
\hline
Dermatology & 10
& $0.699\pm0.160$ & $0.743\pm0.104$ & $0.795\pm0.080$ \\
Car Evaluation & 10
& $0.766\pm0.106$ & $0.800\pm0.088$ & $0.866\pm0.093$ \\
Abalone & 10
& $0.608\pm0.226$ & $0.531\pm0.317$ & $0.617\pm0.161$ \\
Adult & 10
& $0.667\pm0.246$ & $0.685\pm0.137$ & $0.744\pm0.054$ \\
Diabetic & 5
& $0.589\pm0.178$ & $0.684\pm0.090$ & $0.556\pm0.187$ \\
Census-Income & 5
& $0.648\pm0.199$ & $0.708\pm0.106$ & $0.716\pm0.077$ \\
CoverType & 5
& $0.960\pm0.019$ & $0.952\pm0.016$ & $0.958\pm0.024$ \\
\hline
IMDb & 5
& $0.441\pm0.225$ & $0.562\pm0.241$ & $0.507\pm0.266$ \\
TweetEval Sentiment & 5
& $0.169\pm0.113$ & $0.153\pm0.146$ & $0.268\pm0.219$ \\
\hline
CIFAR-10 & 3 & $0.892 \pm 0.049$ & $0.913 \pm 0.043$ & $0.892 \pm 0.042$ \\
DermaMNIST & 3 & $0.476 \pm 0.287$ & $0.553 \pm 0.157$ & $0.533 \pm 0.106$ \\
\hline
\end{tabular}
\end{table}

\subsection{Sensitivity to the Size-Penalty Coefficient}
\label{app:alpha_sensitivity}

\begin{table}[!htbp]
\centering
\footnotesize
\setlength{\tabcolsep}{4pt}
\renewcommand{\arraystretch}{0.95}
\caption{Sensitivity to the size-penalty coefficient.}
\label{tab:alpha_sensitivity}
\begin{tabular}{llccc}
\hline
Dataset & $\alpha$ & Anchor $\rho$ & Held-out $\rho$
& Retained (\%) \\
\hline
Adult & 0
& $0.847 \pm 0.006$ & $0.867 \pm 0.067$ & $25.86 \pm 0.99$ \\
& 0.10
& $0.833 \pm 0.075$ & $0.811 \pm 0.017$ & $19.35 \pm 5.42$ \\
& 0.20
& $0.842 \pm 0.025$ & $0.748 \pm 0.177$ & $22.36 \pm 5.69$ \\
& 0.40
& $0.850 \pm 0.033$ & $0.787 \pm 0.017$ & $20.41 \pm 3.80$ \\
\hline
IMDb & 0
& $0.808 \pm 0.070$ & $0.581 \pm 0.209$ & $27.87 \pm 0.63$ \\
& 0.10
& $0.768 \pm 0.045$ & $0.456 \pm 0.376$ & $23.94 \pm 7.19$ \\
& 0.20
& $0.771 \pm 0.051$ & $0.473 \pm 0.391$ & $23.49 \pm 6.75$ \\
& 0.40
& $0.830 \pm 0.005$ & $0.459 \pm 0.227$ & $21.74 \pm 1.40$ \\
\hline
\end{tabular}

\vspace{2pt}
\begin{minipage}{\linewidth}
\scriptsize
Note: Mean $\pm$ sample SD across three seeds per dataset.
Correlations compare full-data and coreset rankings;
retained (\%) is the percentage of training instances selected.
\end{minipage}
\end{table}

\subsection{Agreement and Score Error on Anchor Configurations}
\label{app:anchor_diagnostics}

\begin{table}[htbp]
\centering
\scriptsize
\caption{Pairwise agreement on the anchor configurations across
subset-selection methods. Higher values indicate stronger agreement with the full-data hyperparameter ranking.}
\label{tab:appendix_agreement}
\begin{tabular}{lccc}
\hline
\textbf{Dataset} & \textbf{RPC} & \textbf{Random} & \textbf{Gradient-Based} \\
\hline
Dermatology & $0.860 \pm 0.018$ & $0.767 \pm 0.066$ & -- \\
Car Evaluation & $0.906 \pm 0.037$ & $0.765 \pm 0.080$ & -- \\
Abalone & $0.843 \pm 0.038$ & $0.666 \pm 0.059$ & -- \\
Adult & $0.853 \pm 0.025$ & $0.736 \pm 0.048$ & -- \\
Diabetes & $0.888 \pm 0.017$ & $0.821 \pm 0.034$ & -- \\
Census-Income & $0.878 \pm 0.033$ & $0.829 \pm 0.037$ & -- \\
CoverType & $0.951 \pm 0.018$ & $0.907 \pm 0.030$ & -- \\
\hline
IMDb & $0.807 \pm 0.018$ & $0.675 \pm 0.086$ & $0.543 \pm 0.075$ \\
TweetEval  & $0.664 \pm 0.070$ & $0.601 \pm 0.089$ & $0.590 \pm 0.040$ \\
CIFAR-10 & $0.953 \pm 0.009$ & $0.937 \pm 0.014$ & $0.914 \pm 0.013$ \\
DermaMNIST & $0.782 \pm 0.082$ & $0.736 \pm 0.105$ & $0.693 \pm 0.074$ \\
\hline
\end{tabular}
\end{table}

\begin{table}[!htbp]
\centering
\footnotesize
\setlength{\tabcolsep}{4pt}
\renewcommand{\arraystretch}{0.95}
\caption{Mean absolute error (MAE) between full-data and subset-based validation performance scores on the anchor configurations. Lower values indicate closer agreement with the full-data performance vector.}
\label{tab:appendix_mae}
\begin{tabular}{lccc}
\hline
\textbf{Dataset} & \textbf{RPC} & \textbf{Random} & \textbf{Gradient-Based} \\
\hline
Dermatology & $0.178 \pm 0.055$ & $0.165 \pm 0.034$ & -- \\
Car Evaluation & $0.092 \pm 0.026$ & $0.079 \pm 0.016$ & -- \\
Abalone & $0.012 \pm 0.003$ & $0.018 \pm 0.003$ & -- \\
Adult & $0.012 \pm 0.005$ & $0.014 \pm 0.003$ & -- \\
Diabetes & $0.008 \pm 0.001$ & $0.009 \pm 0.001$ & -- \\
Census-Income & $0.003 \pm 0.001$ & $0.004 \pm 0.001$ & -- \\
CoverType & $0.065 \pm 0.009$ & $0.065 \pm 0.009$ & -- \\
\hline
IMDb & $0.041 \pm 0.021$ & $0.038 \pm 0.020$ & $0.081 \pm 0.020$ \\
TweetEval  & $0.059 \pm 0.013$ & $0.059 \pm 0.014$ & $0.074 \pm 0.017$ \\
CIFAR-10 & $0.189 \pm 0.030$ & $0.190 \pm 0.035$ & $0.261 \pm 0.037$ \\
DermaMNIST & $0.125 \pm 0.057$ & $0.123 \pm 0.052$ & $0.152 \pm 0.029$ \\
\hline
\end{tabular}

\vspace{2mm}
\footnotesize{\textit{Note:} RPC denotes the proposed ranking-preserving coreset. Gradient-based results are reported only for differentiable text and image classification models and are therefore not applicable to Decision Tree tabular experiments.}
\end{table}

\clearpage
\section{Optimizer-Level Break-Even Analysis}
\label{app:optimizer_breakeven}

\begin{table}[!htbp]
\centering
\footnotesize
\setlength{\tabcolsep}{4pt}
\renewcommand{\arraystretch}{0.8}

\caption{Optimizer-level break-even analysis based on Eq.~(13), reported separately for GA, CHC, and PSO. $T_{\mathrm{sample}}$ denotes the one-time coreset construction time (s), while $T_{\mathrm{full}}$ and $T_{\mathrm{core}}$ denote the per-optimizer full-data and coreset-based HPO runtimes (s), respectively. $N_{\mathrm{BE}}$ represents the number of repeated runs of the corresponding optimizer required to recover the initial $T_{\mathrm{sample}}$ cost. This analysis complements the cycle-level
break-even analysis in Table~\ref{tab:end_to_end_breakeven}, which considers one complete
GA--CHC--PSO optimization cycle.}

\label{tab:optimizer_level_breakeven}
\begin{tabular}{llccccc}
\hline
\textbf{Dataset} & \textbf{Optimizer} & $T_{\text{sample}}$ & $T_{\text{full}}$ & $T_{\text{core}}$ & $T_{\text{sample}}+T_{\text{core}}$ & $N_{\text{BE}}$ \\
\hline
Dermatology & GA & $5.67$ & $0.27$ & $0.23$ & $5.91$ & $157.64$ \\
 & CHC & $5.67$ & $0.74$ & $0.65$ & $6.33$ & $62.36$ \\
 & PSO & $5.67$ & $0.27$ & $0.24$ & $5.92$ & $177.34$ \\
 \hline
Car Evaluation & GA & $6.25$ & $0.37$ & $0.30$ & $6.54$ & $83.28$ \\
 & CHC & $6.25$ & $0.93$ & $0.68$ & $6.92$ & $24.88$ \\
 & PSO & $6.25$ & $0.40$ & $0.34$ & $6.58$ & $94.64$ \\
\hline
Abalone & GA & $9.52$ & $1.92$ & $0.66$ & $10.17$ & $7.52$ \\
 & CHC & $9.52$ & $3.41$ & $1.35$ & $10.87$ & $4.62$ \\
 & PSO & $9.52$ & $1.58$ & $0.50$ & $10.02$ & $8.87$ \\
\hline
Adult & GA & $26.75$ & $10.71$ & $1.91$ & $28.65$ & $3.04$ \\
 & CHC & $26.75$ & $28.48$ & $5.75$ & $32.50$ & $1.18$ \\
 & PSO & $26.75$ & $13.76$ & $2.30$ & $29.05$ & $2.33$ \\
\hline
Diabetes & GA & $890.76$ & $710.68$ & $90.99$ & $981.75$ & $1.44$ \\
 & CHC & $890.76$ & $1608.11$ & $199.28$ & $1090.04$ & $0.63$ \\
 & PSO & $890.76$ & $937.92$ & $114.48$ & $1005.24$ & $1.08$ \\
\hline
Census-Income & GA & $97.56$ & $43.07$ & $9.46$ & $107.01$ & $2.90$ \\
 & CHC & $97.56$ & $128.67$ & $20.27$ & $117.82$ & $0.90$ \\
 & PSO & $97.56$ & $58.75$ & $10.59$ & $108.15$ & $2.03$ \\
\hline
CoverType & GA & $567.70$ & $294.63$ & $73.37$ & $641.07$ & $2.57$ \\
 & CHC & $567.70$ & $1121.62$ & $172.79$ & $740.49$ & $0.60$ \\
 & PSO & $567.70$ & $507.25$ & $88.97$ & $656.67$ & $1.36$ \\
\hline
IMDb & GA & $3427.26$ & $829.13$ & $182.51$ & $3609.76$ & $5.30$ \\
 & CHC & $3427.26$ & $1356.48$ & $550.01$ & $3977.27$ & $4.25$ \\
 & PSO & $3427.26$ & $907.00$ & $273.43$ & $3700.69$ & $5.41$ \\
\hline
TweetEval & GA & $1064.54$ & $330.29$ & $69.75$ & $1134.29$ & $4.09$ \\
 & CHC & $1064.54$ & $1113.40$ & $248.78$ & $1313.32$ & $1.23$ \\
 & PSO & $1064.54$ & $349.81$ & $106.78$ & $1171.32$ & $4.38$ \\
\hline
CIFAR-10 & GA & $13229.94$ & $2559.70$ & $467.03$ & $13696.97$ & $6.32$ \\
 & CHC & $13229.94$ & $4091.20$ & $1262.19$ & $14492.13$ & $4.68$ \\
 & PSO & $13229.94$ & $2767.52$ & $502.61$ & $13732.54$ & $5.84$ \\
\hline
DermaMNIST & GA & $7854.88$ & $1012.29$ & $342.06$ & $8196.94$ & $11.72$ \\
 & CHC & $7854.88$ & $1807.70$ & $438.24$ & $8293.12$ & $5.74$ \\
 & PSO & $7854.88$ & $1327.84$ & $248.14$ & $8103.01$ & $7.28$ \\
\hline
\end{tabular}
\end{table}

\clearpage

\bibliography{sample}

\end{document}